\pdfoutput=1

\documentclass[10pt,sigconf]{acmart}

\usepackage{caption}
\usepackage{setspace}
\usepackage{enumitem}

\usepackage{subcaption}
\usepackage{hyperref}
\usepackage{bbm}
\usepackage{algorithm}
\usepackage{algorithmic}
\usepackage{tabularx}
\usepackage{multirow}

\usepackage{booktabs,subcaption,amsfonts,dcolumn}
\newcolumntype{d}[1]{D..{#1}}

\usepackage{diagbox}
\usepackage{flushend}

\usepackage{mathtools}
\DeclarePairedDelimiter{\ceil}{\lceil}{\rceil}

\DeclareMathOperator*{\argmin}{arg\,min}

\newcommand{\Mod}[1]{\ (\mathrm{mod}\ #1)}
\def \ie {\emph{i.e.}, }

\begin{document}

\title[Understanding and Optimizing Neural Network Execution Time]{FastDeepIoT: Towards Understanding and Optimizing Neural Network Execution Time on Mobile and Embedded Devices}

\acmConference[SenSys]{The 16th ACM Conference on Embedded Networked Sensor Systems}{November 4-7, 2018}{Shenzhen, China}

\author{Shuochao Yao}
\affiliation{%
  \institution{University of Illinois Urbana Champaign}
  }
  
\author{Yiran Zhao}
\affiliation{%
  \institution{University of Illinois Urbana Champaign}
  } 
 
\author{Huajie Shao}
\affiliation{%
 \institution{University of Illinois Urbana Champaign}
  }
  
\author{ShengZhong Liu}
\affiliation{%
 \institution{University of Illinois Urbana Champaign}
  }
  
\author{Dongxin Liu}
\affiliation{%
 \institution{University of Illinois Urbana Champaign}
  }
  
\author{Lu Su}
\affiliation{%
  \institution{State University of New York at Buffalo}
  }

\author{Tarek Abdelzaher}
\affiliation{%
  \institution{University of Illinois Urbana Champaign}
  }

\sloppy

\begin{abstract}
Deep neural networks show great potential as solutions to many sensing application problems, but their excessive resource demand slows down execution time, pausing a serious impediment to deployment on low-end devices. To address this challenge, recent literature focused on compressing neural network size to improve performance. We show that changing neural network size does not proportionally affect performance attributes of interest, such as execution time. Rather, extreme run-time nonlinearities exist over the network configuration space. Hence, we propose a novel framework, called FastDeepIoT, that uncovers the non-linear relation between neural network structure and execution time, then exploits that understanding to find network configurations that significantly improve the trade-off between execution time and accuracy on mobile and embedded devices. FastDeepIoT makes two key contributions. First, FastDeepIoT automatically learns an accurate and highly interpretable execution time model for deep neural networks on the target device. This is done without prior knowledge of either the hardware specifications or the detailed implementation of the used deep learning library. Second, FastDeepIoT informs a compression algorithm how to minimize execution time on the profiled device without impacting accuracy. We evaluate FastDeepIoT using three different sensing-related tasks on two mobile devices: Nexus 5 and Galaxy Nexus. FastDeepIoT further reduces the neural network execution time by $48\%$ to $78\%$  and energy consumption by $37\%$ to $69\%$ compared with the state-of-the-art compression algorithms.
\end{abstract}


\copyrightyear{2018} 
\acmYear{2018} 
\setcopyright{acmcopyright}
\acmConference[SenSys '18]{The 16th ACM Conference on Embedded Networked Sensor Systems}{November 4--7, 2018}{Shenzhen, China}
\acmBooktitle{The 16th ACM Conference on Embedded Networked Sensor Systems (SenSys '18), November 4--7, 2018, Shenzhen, China}
\acmPrice{15.00}
\acmDOI{10.1145/3274783.3274840}
\acmISBN{978-1-4503-5952-8/18/11}


\maketitle

\renewcommand{\shortauthors}{S. Yao et al.}

{

\section{Introduction}
The proliferation of internetworked mobile and embedded devices with growing sensing and computing capabilities promises to revolutionize the interactions between humans and devices that perform complex sensing and recognition tasks.
\begin{table}[!htb]
\vspace{-0.3cm}
\begin{center}
\scriptsize
\caption {Execution time of convolutional layers with $3\times3$ kernel size, stride $1$, same padding, and $224\times 224$ input image size on the Nexus 5 phone.}
\label{tab:flops_exp}
\begin{tabular}{ |c | c | c | c | c | } 
 \hline
 & in\_channel & out\_channel & FLOPs & Time (ms)  \\ 
  \hline
   \hline
  CNN1 &  8 &  32 & 452.4 M & 114.9 \\ 
  \hline
  CNN2 & 32  & 8 & 452.4 M & 300.2 \\ 
 \hline
 CNN3 & 66  & 32 & 3732.3 M & 908.3 \\ 
 \hline
 CNN4 & 43  & 64 & 4863.3 M & 751.7 \\ 
 \hline
\end{tabular}
\end{center}
\vspace{-0.1cm}
\end{table} 
\setlength{\textfloatsep}{0pt}
Much prior work has been dedicated to building smarter and more user-friendly sensing applications in several embedded systems areas, including health and wellness~\cite{xiang2013hybrid,rahman2015dopplesleep,sorber2012plug,bui2017pho2}, context sensing~\cite{wei2015radio,li2013sensor,zhang2016dopenc,nirjon2015typingring,chen2015tracking}, and object detection and localization~\cite{wen2013assessing,he2010listen,langendoen2003distributed,mirshekari2016characterizing,lazik2015alps,eichelberger2017indoor}. 

At the same time, recent advances in deep learning have changed the way computing devices process human-centric content, such as images, speech and audio. Neural network models are especially good at fusing multiple sensing modalities and extracting temporal relationships, which have shown remarkable improvements in audio sensing~\cite{lane2015deepear,georgiev2017low}, tracking and localization~\cite{yao2017deepsense}, human activity recognition~\cite{yao2017deepsense,radu2018multimodal,yao2018sensegan}, and environment sensing~\cite{yao2018rdeepsense}. 
Applying deep neural networks to mobile and embedded devices could thus bring about a generation of applications capable of performing complex sensing and recognition tasks to support a new realm of interactions between humans and their physical surroundings~\cite{yao2018deep}.

The key impediment to wide-spread deployment of deep-learning-based sensing applications remains their high execution time and energy consumption on mobile and embedded devices. Minimizing the execution time of deep neural networks is critical to preserve the real-time properties of such embedded sensing applications as image recognition and object detection in self-driving cars~\cite{Girshick2015FastR, ren2015faster}. 
One promising solution is to compress neural networks into more succinct structures. Traditionally, speeding up neural network execution time is accomplished by reducing the size of model parameters~\cite{han2015deep,yao2017deepiot}. 
Most manually designed time-efficient neural network structures for mobile devices use parameter size or FLOPs as the indicator of execution time~\cite{Zhang2017ShuffleNetAE,Howard2017MobileNetsEC,Iandola2016SqueezeNetAA}. Even the official TensorFlow website recommends to use the total number of floating number operations (FLOPs) of neural networks ``to make rule-of-thumb estimates of how fast they will run on different devices".\footnote{\url{https://www.tensorflow.org/versions/r1.5/mobile/optimizing}}

Although significant progress has been made on neural network structure compression to reduce the resource demands, changing neural network structure has a non-linear effect on system performance, opening opportunities for further performance improvements should such nonlinearities be explicitly identified and exploited. 

\begin{figure}[!htb]
\vspace{-0.3cm}
\centering
\includegraphics[width=0.6\linewidth]{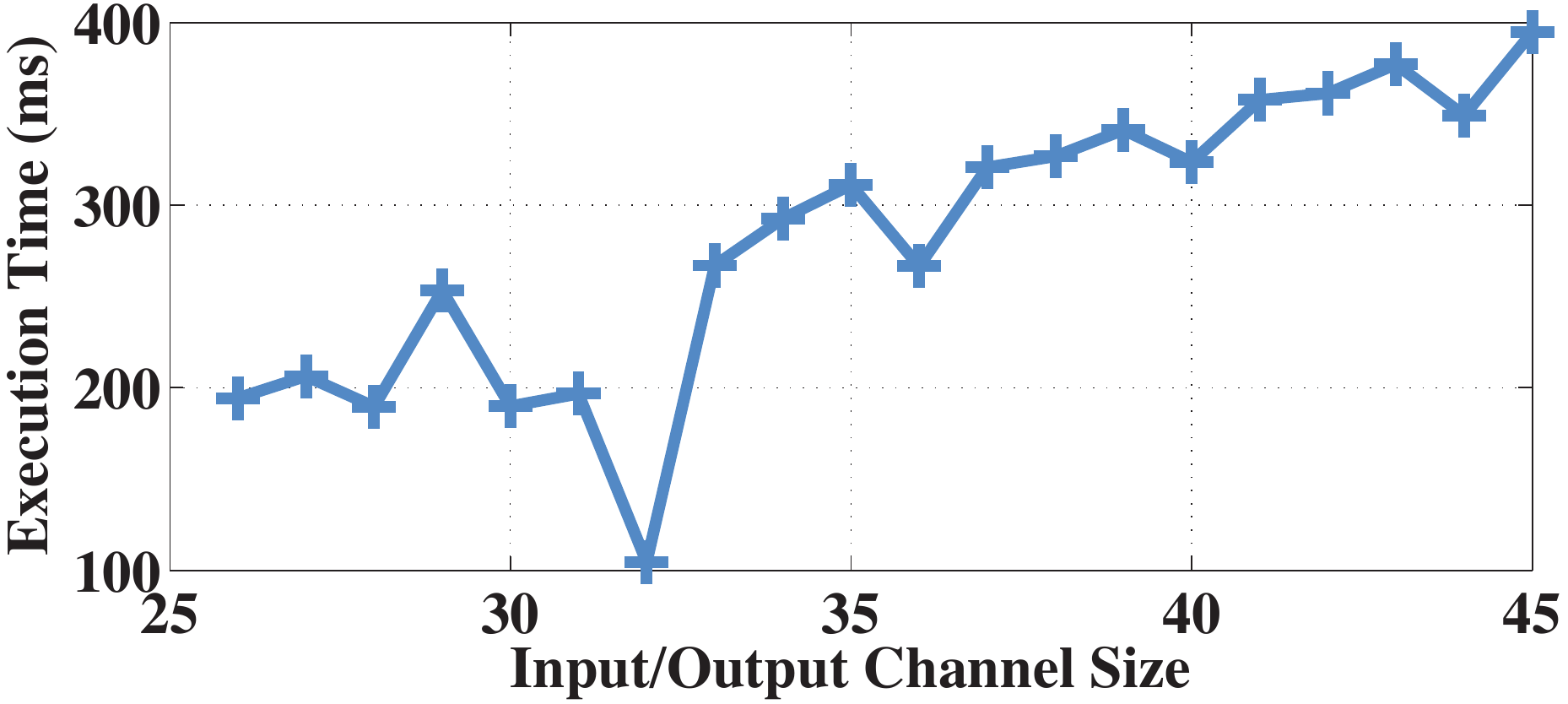}
\caption{The non-linearity of neural network execution time over input/output channel.}
\label{fig:nonlineartiy}
\end{figure}

In this paper, we show how a better understanding of the non-linear relation between neural network structure and performance can further improve execution time and energy consumption without impacting accuracy.
The rest of this paper is organized as follows. The nonlinear relation between network structure and performance is discussed in Section~\ref{sec:nonlinear}. We present the technical details of FastDeepIoT in Section~\ref{sec:model} and  system implementation in Section~\ref{sec:implementation}. The evaluation is presented in Section~\ref{sec:evaluation}. Section~\ref{sec:related} introduces related work. We conclude in Section~\ref{sec:conclusion} introducing avenues for future work.

\vspace{-0.2cm}
\section{Nonlinearities: Evidence and Exploitation}
\label{sec:nonlinear}
In practice, counting the number of neural network parameters and the total FLOPs does not lead to good estimates of execution time because the relation between these predictors and execution time is not proportional. On one hand, 
the fully-connected layer usually has more parameters but takes much less time to run compared to the convolutional layer~\cite{rigamonti2013learning}. On the other hand, one can easily find examples, where increasing the total FLOPs does not translate into added execution time. Caching effects, memory accesses, and compiler optimizations complicate the translation. 
Table~\ref{tab:flops_exp} shows that CNN2 takes around $\times 2.6$ the execution time of CNN1, while both have {\em the same\/} total FLOPs. Moreover, CNN3 takes {\em longer\/} to run compared to CNN4 despite having {\em fewer\/} FLOPs. These observations indicate that current rules-of-thumb for estimating neural network execution time are not the best approximations.

FastDeepIoT answers two key questions to better parameterize neural network implementations for efficient execution on mobile and embedded platforms: 
\begin{enumerate}[topsep=0pt,leftmargin=*]
\setlength\itemsep{0pt}
\item What are the main factors that affect the execution time of neural networks on mobile and embedded devices? 
\item How to guide existing structure
compression algorithms to minimize the neural network execution time properly? 
\end{enumerate}
FastDeepIoT consists of two main modules to tackle these two challenging problems, respectively.

\smallskip
\noindent
{\em Profiling:\/}
Due to different code-level optimizations for different network structures within the deep learning library, the execution time of neural network layers can be extremely nonlinear over the structure configuration space. 
A simple illustration is shown in Figure~\ref{fig:nonlineartiy}, where we plot the execution time of convolutional layers when changing the size of input and output channels simultaneously. 
The plot reveals non-monotonic effects, featuring periodic dips in execution time as network size increases.

A simple regression model over the entire space will thus not be a good approximation. Instead, 
we propose a tree-structured linear regression model. Specifically, we  
automatically detect key conditions at which linearity is violated and arrange them into a tree structure that splits the overall modeling space into piecewise linear regions.
Within each region (tree branch), we use linear regression to convert input structure information into some key explanatory variables, predictive of execution time. The splitting of the overall space and the fitting of subspaces to predictive models are done jointly, which improves both model interpretability and accuracy. 
The aforementioned modeling is done without specific knowledge of  underlying hardware and deep learning library. 

\smallskip
\noindent
{\em Compression:\/} 
Using the results of profiling, we then propose a compression steering module that guides existing neural network structure compression methods to better minimize execution time. 
The execution time model leads compression algorithms to focus more on the layer that takes longer to run instead of treating all layers equally or concentrating on inaccurate total metrics. It is also better able at exploiting non-monotonicity of execution time with respect to network structure size to reduce the former without hurting application-level accuracy metrics. 

We evaluate the profiling and compression steering modules in FastDeepIoT on two devices, Nexus 5 and Galaxy Nexus, with the TensorFlow for Mobile library~\cite{tensorflow_mobile}. 
The profiling module is evaluated on all commonly used network layers, including fully-connected, convolutional, and recurrent layers.
The mean absolute percentage error in estimating execution time is around $1\%$ to $7\%$, which outperforms other complex regression models in most cases.
The compression steering module is evaluated with three representative sensing-related tasks, including vision-based interactions and human activity recognition.  Compared to the state-of-the-art compression algorithms, FastDeepIoT can speed up the neural network execution time by an additional $48\%$ to $78\%$, and improve energy consumption by an additional $37\%$ to $69\%$ on all devices without loss of accuracy.

\vspace{-0.2cm}
\section{System Design}~\label{sec:model}
As mentioned above, the contribution of FastDeepIoT lies in two modules; the profiling module and the compression steering module. 
Below, we introduce the technical details of the two modules, respectively.

\vspace{-0.2cm}
\subsection{Profiling Module}~\label{sec:profiling_module}
We separate this module into two parts. The first part generates diverse training structures for profiling. The second part builds an accurate and interpretable model predicting the execution time of deep learning components for the corresponding structure information.

\begin{table}[!htb]
\begin{center}
\scriptsize
\caption {The scope of our structure configuration for fully-connected (FC), convolutional (CNN), and recurrent (RNN) layers.}
\vspace{-0.1cm}
\label{tab:structure}
\begin{tabular}{ |c | c | } 
 \hline
Type & Structure configuration scope  \\ 
  \hline
   \hline
   FC & \textit{in\_dim} $\in [1, 4096]$ $\quad$ \textit{out\_dim} $\in [1, 4096]$ \\
 \hline
  \multirow{ 4}{*}{CNN} &  \textit{in\_height} $\in[24, 225]$ \quad \textit{in\_width} $\in[24, 225]$    \\ 
  & $\textit{kernel\_height}\times \textit{kernel\_width} \in \{ 2\times2, 3\times3, 4\times4, 5\times5, 2\times3 \}$ \\
  & \textit{in\_channel} $\in [1, 256]$ $\enskip$ \textit{out\_channel} $\in [1, 256]$  \\
  & \textit{padding} $\in \{\text{valid}, \text{same}\}$ $\enskip$ \textit{stride} $\in \{1, 2\}$ \\
  \hline
   \multirow{ 2}{*}{RNN} & \textit{in\_dim} $\in [1, 512]$ $\quad$ \textit{out\_dim} $\in [1, 512]$   \\
   & \textit{step} $\in \{8, 10, 15, 20\}$ \\ 
 \hline
\end{tabular}
\end{center}
\vspace{-0.05cm}
\end{table} 

\vspace{-0.2cm}
\subsubsection{Neural Network Profiling} We introduce the basic system settings and the procedure of generating training structures for profiling here.

FastDeepIoT utilizes TensorFlow benchmark tool~\cite{benchmark_tool} to profile the execution time of all deep learning components on the target device. In order to make the profiling results fully reflect the changes on the neural network structures, we fix the frequencies of phone CPUs (processors) to be constants and stop all the power management services that can affect the processor frequency on target devices, such as fixing \textit{mpdecision} on Qualcomm chips. 

The next step is to generate diverse neural network structures for time profiling. As a deep learning component, such as a convolutional layer and recurrent layer, the combinations of its structure design choices can form an extremely huge structure configuration space. Therefore, we can only select a small proportion of structure configurations during our time profiling. The scope of our structure configuration is shown in Table~\ref{tab:structure}, from which the network generation code chooses a random combination. Notice that we do not contain the activation function as the profiling choice, because it only occupies around $1\% \sim 2\%$ execution time of a deep learning component through empirical observations. By eliminating this insignificant configuration, \ie activation\_function $\in \{\text{ReLU}, \text{Tanh}, \text{sigmoid}\}$, we can save the number of profiling components by the factor of 3. Except for some pre-defined cases, such as sigmoid activation function for gate outputs in recurrent layers, we set all activation functions to be ReLU, which is one of the most widely used activation functions. In addition, the order of deep learning components in the network has little impact on their execution time empirically.

In our profiling module, for each target device, we profile around 120 neural networks with about 1300 deep learning components in total. These time profiling results form a time profiling dataset, $\mathcal{D} = \{\mathcal{S}_i, y_i\}$, where $\mathcal{S}_i$ is the structure configuration and $y_i$ the execution time.

\subsubsection{Execution Time Model Building}~\label{sec:build_run_time_model}
Due to the code-level optimization for different component configuration choices in the deep learning library, execution-time non-linearity appears over the structure configuration space as shown in Figure~\ref{fig:nonlineartiy}. 
The main challenge here is to build a model that can automatically figure out the conditions that cause the execution-time non-linearity without specific knowledge of underlying library and hardware. 

\begin{table*}[!htb]
\vspace{-0.2cm}
\begin{center}
\scriptsize
\caption {The definition of parameter and memory information for Fully-Connected layer (FC), Convolutional layer (CNN), Gated Recurrent Unit (GRU), and Long Short Term Memory (LSTM).}
\label{tab:feature}
\vspace{-0.1cm}
\begin{tabular}{ |c | c | c | c | c | } 
 \hline
Type & \textit{param\_size} & \textit{mem\_in} & \textit{mem\_out} & \textit{mem\_inter}  \\ 
  \hline
   \hline
   FC & $\textit{in\_dim}\times \textit{out\_dim} + \textit{out\_dim}$ & \textit{in\_dim} & \textit{out\_dim} & $0$  \\
 \hline
  \multirow{ 2}{*}{CNN} & $\textit{kernel\_height} \times \textit{kernel\_width} \times$  & $\textit{in\_height} \times \textit{in\_width} \times$ & $\textit{out\_height} \times \textit{out\_width} \times$ &  $\textit{out\_height} \times \textit{out\_width} \times \textit{kernel\_height} \times$ \\
  
 & $\textit{in\_channel} \times \textit{out\_channel}$ +1 & $\textit{in\_channel}$ & $\textit{out\_channel}$ & $\textit{kernel\_width} \times \textit{in\_channel}$ \\
 \hline
 \multirow{ 2}{*}{GRU} & $3\times\textit{out\_dim}\times$ & \multirow{ 2}{*}{$\textit{step} \times \textit{in\_dim}$} & \multirow{ 2}{*}{$\textit{step} \times \textit{out\_dim} $} & \multirow{ 2}{*}{$3\times\textit{step} \times \textit{out\_dim} $}  \\
 & $(\textit{in\_dim} + \textit{out\_dim} + 1)$ & & & \\
  \hline
  \multirow{ 2}{*}{LSTM} & $4\times\textit{out\_dim}\times$ & \multirow{ 2}{*}{$2\times\textit{step} \times \textit{in\_dim}$} & \multirow{ 2}{*}{$2\times\textit{step} \times \textit{out\_dim} $} & \multirow{ 2}{*}{$4\times\textit{step} \times \textit{out\_dim} $} \\
   & $(\textit{in\_dim} + \textit{out\_dim} + 1)$ & & & \\
  \hline
\end{tabular}
\end{center}
\vspace{-0.25cm}
\end{table*} 
\setlength{\textfloatsep}{0pt}

In order to maintain both the accuracy and interpretability, we propose a tree-structure linear regression model.
The model can recursively 
partition the structure configuration space such that the time profiling samples fitting the same linear relationship are grouped together.
The intuition behind this model is
that the execution time of deep learning component under each particular code-level optimization can be formulated with a linear relationship given a set of well-designed explanatory variables.
In addition, different deep learning components, \ie fully connected, convolutional, and recurrent layer, learn their own execution time models.

Each time profiling data is composed of three elements. The feature vector $\mathbf{f}$, used for identifying the condition that causes the execution-time non-linearity; the execution time $y$; and the explanatory variable vector $\mathbf{x}$, used for fitting the execution time $y$.

The basic idea of tree-structure linear regression is to find out the most significant condition causing the execution-time non-linearity within the current dataset recursively. These conditions will form a binary tree structure.
In order to figure out key conditions causing the execution-time non-linearity, we take two conditioning functions into account.
\begin{enumerate}[leftmargin=*,topsep=0pt]
\item \textit{Range condition} $C_1(\mathbf{f}[j],  \tau) := {f}[j] \le  \tau$: identifies execution-time non-linearity caused by cache and memory hit as well as specific implementation for a certain feature range.
\item \textit{Integer multiple condition} $C_2(\mathbf{f}[j],  \tau) := {f}[j]  \equiv 0 \Mod{\tau}$: identifies execution-time non-linearity caused by loop unrolling, data alignment, and parallelized operations. 
\end{enumerate}

Assume that we are generating node $m$ in the binary tree with dataset $\mathcal{D}_m$. The model creates a set of conditions $\{\phi\}$. Each of them can partition the dataset into two subsets $\mathcal{D}_m^{(l)}(\phi)$ and $\mathcal{D}_m^{(r)}(\phi)$. Each condition $\phi$ consists of three elements, $\phi = \{ \mathbf{f}[j],  \tau_m, k\}$, where $k \in \{1,2\}$ is the conditioning function type.
\vspace{-0.3cm}
\begin{equation}
\small
\begin{split}
\mathcal{D}_m^{(l)}(\phi) & =  \mathcal{D}_m| C_k(x_j,  \tau_m), \\
\mathcal{D}_m^{(r)}(\phi) & = \mathcal{D}_m \textbackslash \mathcal{D}_m^{(l)}(\phi).
\end{split}
\label{eqn:partition}
\end{equation}

Node $m$ selects the most significant condition $\phi^*$ by minimizing the impurity function $G(\mathcal{D}_m, \phi)$,

\vspace{-0.3cm}
\begin{small}
\begin{eqnarray}
\phi^* & =&  \argmin_{\phi} G(\mathcal{D}_m, \phi),~\label{eqn:min_impurity}\\
G(\mathcal{D}_m, \phi) &= &\frac{\big\vert \mathcal{D}_m^{(l)}(\phi) \big\vert}{\vert \mathcal{D}_m \vert} H\big(\mathcal{D}_m^{(l)}(\phi) \big) + \frac{\big\vert \mathcal{D}_m^{(r)}(\phi) \big\vert}{\vert \mathcal{D}_m \vert}  H\big(\mathcal{D}_m^{(r)}(\phi) \big),~\label{eqn:impurity_fun}\\
H(\mathcal{D}) &=& \min_{\mathbf{w}, b} \frac{1}{\vert \mathcal{D}\vert} \sum_{(\mathbf{x}, y) \in \mathcal{D}} (\mathbf{w}^\intercal \mathbf{x} + b - y)^2 \quad \text{s.t.} \phantom{a} \mathbf{w}, b \ge 0.
~\label{eqn:line_regression}
\end{eqnarray}
\end{small}

\begin{algorithm}        
\scriptsize
\caption{Execution time model building}   
\label{alg:run_time}                     
\begin{algorithmic}[1]                
\STATE \textbf{Input:} time profiling dataset $\mathcal{D}_z$, feature vector $\mathbf{f}$, two conditioning functions ${f}[j] \le  \tau$ and ${f}[j]  \equiv 0 \Mod{\tau}$, and explanatory variable $\mathbf{x}_{z}$.
\STATE Fit $\mathcal{D}_z$ with $\mathbf{w}_{r}$ and $b_{r}$ according to~\eqref{eqn:line_regression}.
\STATE Save root node $r = [\emptyset, \mathbf{w}_{r}, b_{r}]$
\STATE \textbf{Initialize:} $\text{que}= \big[[r, \mathcal{D}_z ]\big]$.
\WHILE {len$(\text{que}) > 0$}
\STATE $q, \mathcal{D}_q = \text{que.deque}()$
\IF{$\mathcal{D}_q$ meets stoping condition}
\STATE Continue
\ENDIF
\STATE Search for the optimal partition $\phi^* = \{f[j^*], \tau^*, k^*\}$ according to~\eqref{eqn:min_impurity}~\eqref{eqn:impurity_fun}~\eqref{eqn:line_regression}.
\STATE Generate partitioned dataset $\mathcal{D}_q^{(l)}(\phi^*)$ and $\mathcal{D}_m^{(r)}(\phi^*)$ according to~\eqref{eqn:partition}.
\STATE Fit $\mathcal{D}_q^{(l)}(\phi^*)$ with $\mathbf{w}_{q}^{(l)}$ and $b_{q}^{(l)}$ according to~\eqref{eqn:line_regression}.
\STATE Fit $\mathcal{D}_q^{(r)}(\phi^*)$ with $\mathbf{w}_{q}^{(r)}$ and $b_{q}^{(r)}$ according to~\eqref{eqn:line_regression}.
\STATE Save $q$'s left child node $q^{(l)} = \big[[\text{True}, \phi^*], \mathbf{w}_{q}^{(l)}, b_{q}^{(l)}\big]$
\STATE Save $q$'s right child node $q^{(r)} = \big[[\text{False}, \phi^*], \mathbf{w}_{q}^{(r)}, b_{q}^{(r)}\big]$
\STATE $\text{que.enque}\big([q^{(l)}, \mathcal{D}_q^{(l)}(\phi^*)]\big)$
\STATE $\text{que.enque}\big([q^{(r)}, \mathcal{D}_q^{(r)}(\phi^*)]\big)$
\ENDWHILE
 \end{algorithmic}
\end{algorithm}

\noindent
The impurity function is designed as the weighted mean square errors of linear regressions over two sub-datasets partitioned by the condition $\phi$.

Next, we describe the feature vector $\mathbf{f}$.
Our choice of feature vector $\mathbf{f}$ contains three parts: the structure features, the memory features, and the parameter feature. The structure features refer to \textit{in\_dim} and \textit{out\_dim} for fully-connected and recurrent layers as well as \textit{in\_channel} and \textit{out\_channel} for convolutional layers. The memory features include the memory size of input, \textit{mem\_in}, the memory size of output, \textit{mem\_out}, and the memory size of internal representations, \textit{mem\_inter}. The parameter feature refers to the size of parameters, \textit{param\_size}. The detailed definitions of memory and parameter features are shown in Table~\ref{tab:feature}. All notations in Table~\ref{tab:feature} are consistent with the notations of structure configurations in Table~\ref{tab:structure}, except for the height and width of output image, \textit{out\_height} and \textit{out\_width}, in the convolutional layer. 
However, we can easily calculate these two values based on other structure information, \ie \textit{in\_height}, \textit{in\_width}, \textit{kernel\_height}, \textit{kernel\_width}, \textit{stride}, and \textit{padding}~\footnote{\url{https://www.tensorflow.org/api_guides/python/nn\#Convolution}}.

Last, we discuss about our explanatory variable vector $\mathbf{x}$ for linear regression. In this paper, we build an intuitive performance model that the execution time of a program is contributed by three parts, CPU operations, memory operations, and disk I/O operations. For a deep learning component, these  parts  refer to FLOPs, memory size, and parameter size,
\begin{equation}
\mathbf{x} = [\text{FLOPs}, mem, param\_size].
\label{eqn:predictive}
\end{equation}
where $mem = mem\_in + mem\_out + mem\_inter$. 

With the weight vector $\mathbf{w}$ and the bias term $b$, the overall execution time of a deep learning component, $y$, can be modelled as $y = \mathbf{w}^\intercal \mathbf{x} + b$. Since every term should have a positive contribution to the execution time, we add an additional constraint, $\mathbf{w}, b \ge 0$,  as shown in~\eqref{eqn:line_regression}.

\begin{table}[!htb]
\vspace{-0.2cm}
\begin{center}
\scriptsize
\caption {The p-values of explanatory variables.}
\label{tab:p_values}
\vspace{-0.1cm}
\begin{tabular}{ |c | c | c | c | c | } 
 \hline
Type & FLOPs & \textit{mem} & \textit{param\_size} & \textit{step}  \\ 
  \hline
   \hline
   FC & $0.000$ & $0.000$ & $1.000$ &  \diagbox[dir=SW,width=1.1cm, height=0.25cm]{}{} \\
  \hline
  CNN & $0.037$ & $0.009$ & $1.000$ &   \diagbox[dir=SW,width=1.1cm, height=0.25cm]{}{} \\
  \hline
  GRU & $0.000$ & $0.000$ & $1.000$ & $0.000$  \\
  \hline
  LSTM & $0.000$ & $0.000$ & $1.000$ & $0.000$  \\
  \hline
\end{tabular}
\end{center}
\vspace{-0.05cm}
\end{table}
\setlength{\textfloatsep}{0pt}

The tree-structure linear regression model builds a binary tree that gradually picks out conditions that cause execution-time non-linearity and breaks the dataset into subsets that contain more ``linearity". Our designed explanatory variable vector $\mathbf{x}$ is able to fit the dataset with linear relationships better level by level, especially for fully-connected and convolutional layer.
The recurrent layers, however, still have flaws. We analyze the error and find out that recurrent layers have a constant initialization overhead or set-up time for each step. Therefore, we update explanatory variable vector $\mathbf{x}$,
\begin{equation}
\begin{split}
\mathbf{x}_{fc} &= \mathbf{x}_{cnn} = [\text{FLOPs}, mem, param\_size], \\
\mathbf{x}_{rnn} &= [\text{FLOPs}, mem, param\_size, step]. 
\end{split}
\label{eqn:predictive_updated}
\end{equation}

\begin{figure*}[!htb]
\vspace{-0.3cm}
\begin{minipage}[b]{125mm}
\centering
\includegraphics[width=0.8\linewidth]{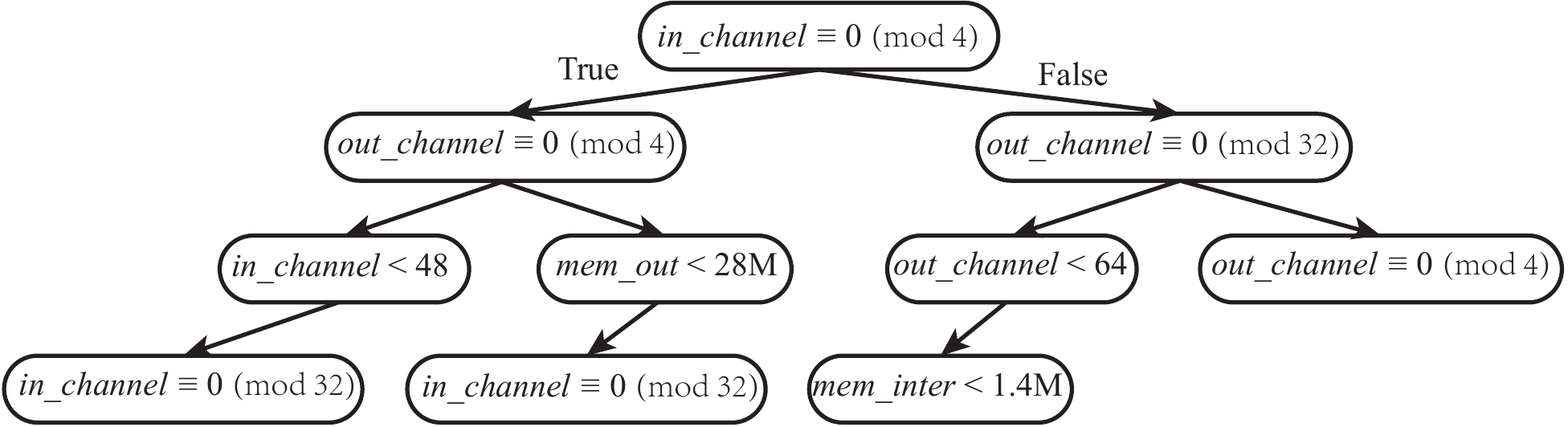}
\vspace{-0.1cm}
\caption{The execution time model of convolutional layers on Nexus 5.}
\label{fig:run_time_nexus5_cnn}
\end{minipage}
\begin{minipage}[b]{50mm}
\centering
\includegraphics[width=0.5\linewidth]{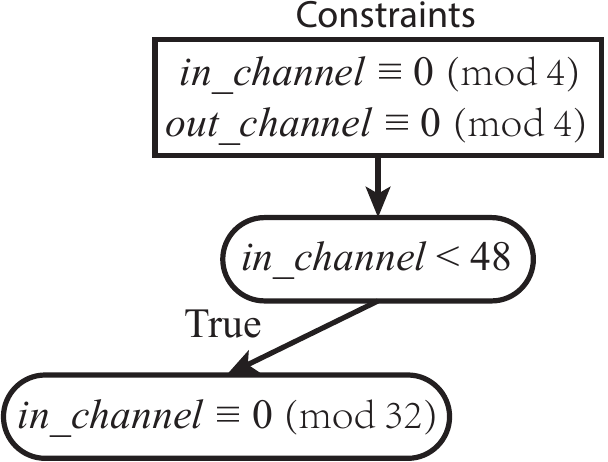}
\vspace{-0.1cm}
\caption{Simplified execution time model of convolutional layers on Nexus 5.}
\label{fig:reduced_run_time_nexus5_cnn}
\end{minipage}
\vspace{-0.3cm}
\end{figure*}

We summarize our execution time model building process in Algorithm~\ref{alg:run_time}. There is a stopping condition in Line 7 that keeps tree-structure linear regression from growing infinitely. 
In our case, the stopping condition occurs when a linear regression can fit the current dataset $\mathcal{D}_q$ with a mean absolute percentage error less than 5\% or when the size of current dataset is smaller than 15, $|\mathcal{D}_q| < 15$.

\vspace{-0.2cm}
\subsubsection{Execution Time Model with Statistical Analysis}
In this part, we provide an illustration of the FastDeepIoT profiling module on Nexus 5 phone with statistical analysis. The module first profiles and generates the execution time profiling dataset. 
Then, the module builds an execution time model for each deep learning component based on the tree-structure linear regression in Algorithm~\ref{alg:run_time}. 
Additional evaluations on the execution time model will be shown in Section~\ref{sec:exe_time_exp}. 

For fully-connected layers and recurrent layers, including GRU and LSTM, their execution time has a perfect linear relationship with our explanatory variable vector $\mathbf{x}_{fc}$ and $\mathbf{x}_{rnn}$. 
However, the execution time model of convolutional layers reflects a strong non-linearity over the structure configuration space. As shown in Figure~\ref{fig:run_time_nexus5_cnn}, 
the execution time of convolutional layer has local minima when \textit{in\_channel} or \textit{out\_channel} is a multiple of $4$. 

Then we calculate the p-values to evaluate the mathematical relationship between each explanatory variable and the execution time. 
The p-value for each explanatory variable tests the null hypothesis that the variable has no correlation with the execution time. Results are shown in Table~\ref{tab:p_values}. 
The p-values of explanatory variables, FLOPs, \textit{mem}, and \textit{step}, are less than the significance level ($0.05$) for all deep learning components. 
So our empirical time profiling data provides enough evidence  that the correlation between these explanatory variables and the execution time are statistically significant. 
However, the p-values for \textit{param\_size} is high for all cases, which shows that the number of parameters has limited correlation with the execution time. This experiment, again, highlights the importance of proposing a compression algorithm targeting on minimizing the execution time instead the number of parameters.

\vspace{-0.2cm}
\subsection{Compression Steering Module}
\vspace{-0.1cm}
Profiling and modelling deep learning execution time  is not enough for speeding up the model execution. In this section, we introduce the compression steering module that is designed to empower existing deep learning structure compression algorithms to minimize model execution time properly.

We assume that $\mathcal{S} = \{\mathbf{s}_l\}$ and $\mathcal{W} = \{\mathbf{W}_l\}$ for $l = 1,\cdots,L$ is structure information and weight matrix of a neural network from layer $1$ to layer $L$ respectively. We denote our execution time model as $t_l =\mathcal{T}(\mathbf{s}_l)$, which takes the structure information $\mathbf{s}_l$ as input and predicts the component execution time $t_l $. For a general neural network structure compression algorithm, we denote the original compression process as,
\begin{equation}
\small
\min_{\mathcal{S}, \mathcal{W}} \mathcal{L}_\theta(\mathcal{S}, \mathcal{W}), \label{eqn:org_compress}
\end{equation}
where the compression algorithm minimizes a loss function, concerning prediction error or parameter size, with either the gradient descend or searching based optimization method.

In order to enable the compression algorithm to minimize the execution time, our first step is to incorporate the execution time model into the original objective function~\eqref{eqn:org_compress},
\begin{equation}
\small
\min_{\mathcal{S}, \mathcal{W}} \mathcal{L}_\theta(\mathcal{S}, \mathcal{W}) + \lambda \sum_{l=1}^L \mathcal{T}(\mathbf{s}_l), \label{eqn:naive_compress}
\end{equation}
where $\lambda$ is a hyper-parameter that make the tradeoff between minimizing training loss and minimizing execution time.

\begin{algorithm}      
\scriptsize
\caption{Layer expansion and local minima searching}   
\label{alg:local_minima}                     
\begin{algorithmic}[1]                
\STATE \textbf{Input:} the execution time model $\mathcal{T}()$ with root node $r$ and the layer structure $\{\mathbf{f}, \mathbf{x}\}$.
\STATE Set $\text{node} = r$, $\text{condL} = []$.
\WHILE {$\neg(\text{node.left} == \text{None} \enskip \& \enskip \text{node.right} == \text{None})$}
\IF{node.cond is a range condition}
\STATE condL.append(node.cond)
\IF{$\mathbf{f}$ obeys node.cond}
\STATE $\text{node} = \text{node.left}$
\ELSE
\STATE $\text{node} = \text{node.right}$
\ENDIF
\ELSE
\STATE $\hat{\mathbf{f}} = \mathbf{f}$ and $\hat{\mathbf{x}} = \mathbf{x}$.
\STATE $\hat{\mathbf{f}}[\text{node.}j]  = \text{node.}\tau\times\ceil[\big]{\mathbf{f}[\text{node.}j]/\text{node.}\tau} $
\STATE Update $\hat{\mathbf{x}}$ according to $\hat{\mathbf{f}}$.
\IF {$\text{node.}\mathbf{w}_T^\intercal \hat{\mathbf{x}} + \text{node.}b_T > \text{node.}\mathbf{w}_F^\intercal {\mathbf{x}} + \text{node.}b_F  \enskip\&\enskip \hat{\mathbf{f}} \text{ obeys condL}$}
\STATE $\mathbf{f} = \hat{\mathbf{f}}$ and $\mathbf{x} = \hat{\mathbf{x}}$.
\STATE $\text{node} = \text{node.left}$
\ELSE
\STATE $\text{node} = \text{node.right}$
\ENDIF
\ENDIF
\ENDWHILE
\STATE \textbf{Return}: $\mathbf{f}$
 \end{algorithmic}
\end{algorithm}
\setlength{\textfloatsep}{0pt}

Adding execution time to the compression objective function can encourage the compression algorithm to concentrate more on the layers with higher execution time, which helps to speed up the whole neural network. 

However, due to the existence of execution-time local minima, compressing neural network structure is not always the optimal choice for minimizing the execution time. As shown in Figure~\ref{fig:nonlineartiy}, enlarging neural network structure can find a nearby execution-time local minimum that reduces the execution time. Notice that enlarging structure is a lossless operation. We can at least enlarge weight matrices with zeros that keeps performance the same.

In general, utilizing execution-time local minima for speeding up involves two steps:
\begin{enumerate}[topsep=0pt]
\item Identifying an expanded structure configuration that can trigger a nearby execution-time local minimum.
\item Deciding whether the expanded structure can speed up the execution time.
\end{enumerate}

For an execution time model trained with a complex method, such as neural networks, identifying a nearby execution-time local minimum can be almost impossible by blindly searching a large configuration space. However, our tree-structure linear regression can easily identify a nearby local minimum speeding up the neural network execution.
\begin{figure}[!htb]
\vspace{-0.1cm}
\centering
\includegraphics[width=0.6\linewidth]{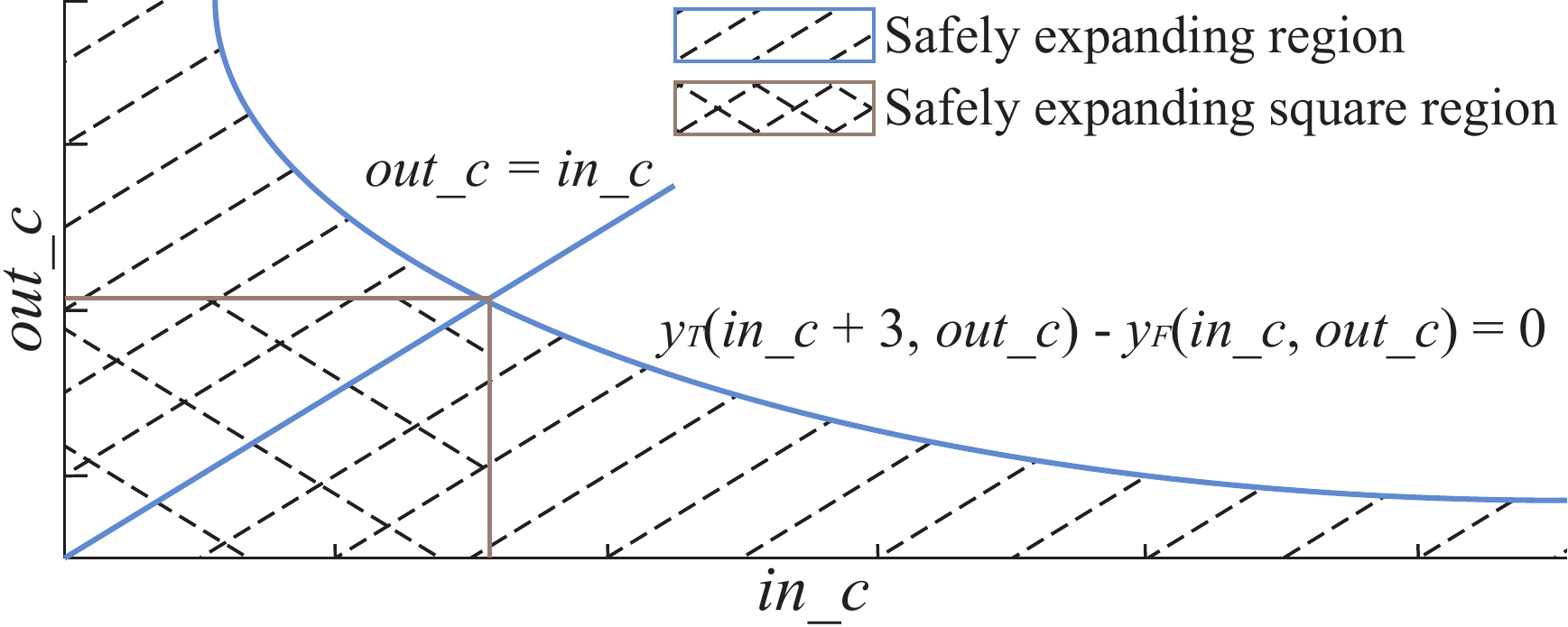}
\vspace{-0.25cm}
\caption{The square region of safely expanding $in\_channel$ for speed up.}
\label{fig:zero_contour}
\end{figure}
\setlength{\textfloatsep}{0pt}
Local extrema, \ie maxima and minima, are identified by the integer multiple condition, ${f}[j]  \equiv 0 \Mod{\tau}$, in our tree-structure linear regression model. Our compression steering module searches for the nearby local maxima by gradually expanding the structure that fits the integer multiple conditions from root node to leaf node in the execution time model.

Assume that node $m$ is under the condition ${\mathbf{f}}[j_m]  \equiv 0 \Mod{\tau_m}$ with two sets of linear regression parameters $\{\mathbf{w}_T, b_T\}$ and $\{\mathbf{w}_F, b_F\}$ used for fitting the dataset that obeying and against the condition respectively. A deep learning layer is denoted with the feature vector $\mathbf{f}_l$ and the explanatory variable vector $\mathbf{x}_l$. The compression steering module generates an expanded layer with feature vector $\hat{\mathbf{f}}_l$ and explanatory variable vector $\hat{\mathbf{x}}_l$ by updating the conditioning feature $\hat{\mathbf{f}}[j_m]  = \tau_m\ceil[\big]{\mathbf{f}[j_m]/\tau_m} $. Then the module compares the values between $\mathbf{w}_T^\intercal \hat{\mathbf{x}} + b_T$ and $\mathbf{w}_F^\intercal {\mathbf{x}} + b_F$ to decide whether it should accept the expansion for speeding-up and go through the corresponding branch.

The layer structure expansion and local minima searching process is summarized in Algorithm~\ref{alg:local_minima}. The algorithm goes through whole tree structure to find out a nearby local minimum that reduces the execution time. 

For a whole neural network, each layer goes through the structure expansion and local minima searching process one by one. It is possible that conflicts exist between expanded structures of two neighbouring layers. The module solves these conflicts sequentially by choosing the one having shorter overall execution time. 

In addition, we can further analyze the structure expansion process for a particular component on a particular device for a particular application settings. For example, assume that we are compressing the \textit{in\_channel} and \textit{out\_channel} of a convolutional layer on Nexus 5 with kernel size $3\times 3$, input image size $24\times 24$, 
and the same padding. We are considering the root condition $\textit{in\_channel}  \equiv 0 \Mod{4}$ as shown in Figure~\ref{fig:run_time_nexus5_cnn}. According to our execution time model, two linear regression models that fit the two datasets in the left and right child of the root node are:
\begin{equation}
\small
\begin{split}
\mathbf{w}_T &= [3.41\times10^{-8},   4.03\times10^{-6},   7.11\times10^{-25}] \enskip b_T = 8.11, \\
\mathbf{w}_F &= [3.11\times10^{-8},   8.03\times10^{-6},   1.52\times10^{-34}] \enskip b_F = 12.82.
\end{split}
\label{eqn:root_lr_weight}
\end{equation}

\noindent
Then we can obtain the execution time as a function of \textit{in\_channel} and \textit{out\_channel} by substituting the explanatory variable vector $\mathbf{x}$ with definitions illustrated in Table~\ref{tab:feature} as well as the application settings about kernel size,  input image size, and padding option. 
\begin{equation}
\small
\begin{split}
y_T(\textit{in\_c}, \textit{out\_c}) =& 3.53\times10^{-4}\cdot\textit{in\_c}\cdot\textit{out\_c} + 8.11  + \\ 
& 2.32\times10^{-2}\cdot\textit{in\_c} + 2.32\times10^{-3}\cdot\textit{out\_c}, \\
y_F(\textit{in\_c}, \textit{out\_c}) =& 3.23\times10^{-4}\cdot\textit{in\_c}\cdot\textit{out\_c} + 12.82 +   \\
&  4.63\times10^{-2}\cdot\textit{in\_c}+ 4.63\times10^{-3}\cdot\textit{out\_c},
\end{split}
\label{eqn:root_run_time}
\end{equation}

\noindent
where we denote \textit{in\_channel} and \textit{out\_channel} as \textit{in\_c} and \textit{out\_c} for simplicity. 

We are interested in the region where expanding the \textit{in\_channel} to a nearby multiple of 4 can speed up the execution. This is equivalent to solving
\begin{equation}
\small
y_T(\textit{in\_c} + 3, \textit{out\_c}) - y_F(\textit{in\_c}, \textit{out\_c}) < 0,
\end{equation}

\noindent
where its zero contour line is a hyperbola. Therefore, within the region bounded by \textit{in\_channel} axis, \textit{out\_channel} axis, and zero contour line, we can safely expand \textit{in\_channel} to a multiple of 4 to speed up the convolutional layer execution time. 

In order to have a more interpretable result, as shown in Figure~\ref{fig:zero_contour}, we can obtain a square region by finding the intersections between the zero contour line and the function $out\_channel = in\_channel$. 
In this case, within the region $in\_channel\times out\_channel \in [1, 1288]\times [1, 1288]$, we can blindly expand \textit{in\_channel} to a multiple of 4 to speed up. This region is much larger than the region we are interested in. We can keep analyzing the next condition $\textit{out\_channel}  \equiv 0 \Mod{4}$ and achieve similar result. Within the region $in\_channel\times out\_channel \in [1, 808]\times [1, 808]$, we can safely expand \textit{in\_channel} and \textit{out\_channel} to a nearby multiple of 4 to speed up. In the end, we can obtain a simplified execution time model $ \hat{\mathcal{T}}$ as shown in Figure~\ref{fig:reduced_run_time_nexus5_cnn}.

In summary, the compression steering module compresses the neural network structure for reducing overall execution time with three steps.
\begin{enumerate}[topsep=0pt,topsep=0pt]
\item Compressing neural network with a time-aware objective function~\eqref{eqn:naive_compress} with execution time model $\mathcal{T}$.
\item Expanding layer structure and searching local minima for further speed up according to Algorithm~\ref{alg:local_minima} with execution time model $\mathcal{T}$ or $\hat{\mathcal{T}}$ (if available).
\item Depending on the original compression algorithm, freeze the structure and fine-tune the neural network.
\end{enumerate}

\vspace{-0.1cm}
\section{Implementation}~\label{sec:implementation}
In this section, we briefly describe the hardware, software, and architecture of FastDeepIoT. 

\vspace{-0.2cm}
\subsection{Hardware}
In this paper, we test FastDeepIoT on two types of hardware, Nexus 5 phone and Galaxy Nexus phone. Two devices are profiled for each type of hardware.
The Nexus 5 phone is equipped with quad-core 2.3 GHz CPU and 2 GB memory. 
The Galaxy Nexus phone is equipped with dual-core 1.2 GHz CPU and 1GB memory. We stop the \textit{mpdecision} service and use \textit{userspace} CPU governor for two hardware. 
We manually set 1.1GHz for the quad-core CPU on Nexus 5, and 700MHz for the dual-core CPU on Galaxy Nexus to prevent overheating caused by the constant time profiling.
In addition, all profiling and testing neural network models are run solely on CPU.
The execution time model building and the compression steering module are implemented on a workstation connected to two phones. 

\vspace{-0.2cm}
\subsection{Software}
FastDeepIoT utilizes TensorFlow benchmark tool~\cite{benchmark_tool}, a C++ binary, to profile the execution time of deep learning components. For each neural network, the benchmark tool have one warm up run to initialize the model and then profile all components execution time with 20 runs without internal delay. Mean values are taken as the profiled execution time. 

We install Android 5.0.1 on Nexus 5 phone and Android 4.3 on Galaxy Nexus phone. All additional background services are closed during the profiling and testing. All energy consumptions are measured by an external power meter. 

\vspace{-0.2cm}
\subsection{Architecture}
Given a target device, FastDeepIoT first queries the device and its own database for a pre-generated execution time model with device type and OS version as the key. If the query fails, the profiling module starts its function. FastDeepIoT generates random neural network structures based on the configuration scope in Table~\ref{tab:structure}, pushes the Protocol Buffers (.pb file) to the target device, profiles the execution time of components, fetches back and processes the profiling result. Once the profiling process has finished, FastDeepIoT learns tree-structure linear regression execution time models according to Algorithm~\ref{alg:run_time} based on the time profiling dataset. FastDeepIoT pushes the generated execution time models to the target device and its own database for storage. 

Then given an original neural network structure and parameters, the compression steering module can automatically generate a compressed structure to speed up inference time for a target device. FastDeepIoT queries the target device and own database for a pre-generated execution time model, and choose a structure compression algorithm, DeepIoT as a default, to reduce the deep learning execution time according to \eqref{eqn:naive_compress} and Algorithm~\ref{alg:local_minima}. The resulting compressed neural network is transferred to the target device used locally.

\vspace{-0.2cm}
\section{Evaluation}~\label{sec:evaluation}
In this section, we evaluate FastDeepIoT through two sets of experiments. The first set evaluates the accuracy of the execution time model generated by our profiling module, while the second set evaluates the performance of our compression steering module. 
\begin{table}
\vspace{-0.1cm}
\caption{The Mean Absolute Percentage Error (MAPE), Mean Absolute Error (MAE) in millisecond, and Coefficient of determination ($\text{R}^2$) of execution time models.}
\vspace{-0.15cm}
\label{tab:time_pred}
\begin{subtable}[t]{0.5\textwidth}
\caption{Nexus 5-Convolutional layer}
\vspace{-0.15cm}
\label{tab:time_pred_cnn}
\centering
\scriptsize
\begin{tabular}{ |c | c | c | c | c | c | c | } 
 \hline
 & FastDeepIoT & SVR & DT & RF & GBRT & DNN  \\ 
  \hline
   \hline
   MAPE & $\mathbf{7.6\%}$ & $233.8\%$ & $23.8\%$ & $19.7\%$ & $10.9\%$ & $\mathbf{6.4\%}$  \\
  \hline
  MAE & $\mathbf{15.2}$ & $227.1$ & $39.2$ & $27.3$ & $20.5$ & $16.4$ \\
  \hline
  $\text{R}^2$ & $\mathbf{0.991}$ & $-0.229$ & $0.969$ & $0.985$ & $0.988$ & $\mathbf{0.994}$ \\
  \hline
\end{tabular}
\end{subtable}

\begin{subtable}[t]{0.5\textwidth}
\caption{Nexus 5-Gated recurrent unit}
\vspace{-0.15cm}
\label{tab:time_pred_gru}
\centering
\scriptsize
\begin{tabular}{ |c | c | c | c | c | c | c | } 
 \hline
 & FastDeepIoT & SVR & DT & RF & GBRT & DNN  \\ 
  \hline
   \hline
   MAPE & $\mathbf{1.8\%}$ & $78.7\%$ & $9.4\%$ & $6.7\%$ & $4.8\%$ & $2.0\%$  \\
  \hline
  MAE & $\mathbf{0.6}$ & $23.6$ & $2.9$ & $1.8$ & $1.5$ & $0.7$ \\
  \hline
  $\text{R}^2$ & $\mathbf{0.999}$ & $-0.078$ & $0.986$ & $0.995$ & $0.995$ & $\mathbf{0.999}$ \\
  \hline
\end{tabular}
\end{subtable}

\begin{subtable}[t]{0.5\textwidth}
\caption{Nexus 5-Long short term memory}
\vspace{-0.15cm}
\label{tab:time_pred_lstm}
\centering
\scriptsize
\begin{tabular}{ |c | c | c | c | c | c | c | } 
 \hline
 & FastDeepIoT & SVR & DT & RF & GBRT & DNN  \\ 
  \hline
   \hline
   MAPE & $\mathbf{2.3\%}$ & $73.7\%$ & $9.0\%$ & $4.7\%$ & $4.1\%$ & $2.8\%$  \\
  \hline
  MAE & $\mathbf{0.6}$ & $23.7$ & $3.0$ & $1.4$ & $1.6$ & $0.9$ \\
  \hline
  $\text{R}^2$ & $\mathbf{0.999}$ & $-0.223$ & $0.977$ & $0.995$ & $0.993$ & $0.998$ \\
  \hline
\end{tabular}
\end{subtable}

\begin{subtable}[t]{0.5\textwidth}
\caption{Nexus 5-Fully-connected layer}
\vspace{-0.15cm}
\label{tab:time_pred_fc}
\centering
\scriptsize
\begin{tabular}{ |c | c | c | c | c | c | c | } 
 \hline
 & FastDeepIoT & SVR & DT & RF & GBRT & DNN  \\ 
  \hline
   \hline
   MAPE & $\mathbf{1.9\%}$ & $133.5\%$ & $22.5\%$ & $12.0\%$ & $\textbf{0.2\%}$ & $1.9\%$  \\
  \hline
  MAE & $\mathbf{0.19}$ & $5.98$ & $1.18$ & $0.38$ & $\mathbf{0.01}$ & $0.19$ \\
  \hline
  $\text{R}^2$ & $\mathbf{0.999}$ & $0.065$ & $0.977$ & $0.996$ & $\mathbf{0.999}$ & $\mathbf{0.999}$ \\
  \hline
\end{tabular}
\end{subtable}
\begin{subtable}[t]{0.5\textwidth}
\caption{Galaxy Nexus-Convolutional layer}
\vspace{-0.15cm}
\label{tab:time_pred_cnn2}
\centering
\scriptsize
\begin{tabular}{ |c | c | c | c | c | c | c | } 
 \hline
 & FastDeepIoT & SVR & DT & RF & GBRT & DNN  \\ 
  \hline
   \hline
   MAPE & $\mathbf{4.1\%}$ & $164.3\%$ & $33.1\%$ & $23.0\%$ & $15.2\%$ & $14.5\%$  \\
  \hline
  MAE & $\mathbf{26.8}$ & $878.7$ & $162.9$ & $123.7$ & $114.6$ & $110.1$ \\
  \hline
  $\text{R}^2$ & $\mathbf{0.999}$ & $-0.246$ & $0.969$ & $0.980$ & $0.982$ & $0.983$ \\
  \hline
\end{tabular}
\end{subtable}

\begin{subtable}[t]{0.5\textwidth}
\caption{Galaxy Nexus-Gated recurrent unit}
\vspace{-0.15cm}
\label{tab:time_pred_gru2}
\centering
\scriptsize
\begin{tabular}{ |c | c | c | c | c | c | c | } 
 \hline
 & FastDeepIoT & SVR & DT & RF & GBRT & DNN  \\ 
  \hline
   \hline
   MAPE & $\mathbf{2.9\%}$ & $71.5\%$ & $10.5\%$ & $8.8\%$ & $6.0\%$ & $4.1\%$  \\
  \hline
  MAE & $\mathbf{1.1}$ & $27.8$ & $4.8$ & $4.1$ & $3.2$ & $2.2$ \\
  \hline
  $\text{R}^2$ & $\mathbf{0.997}$ & $-0.065$ & $0.968$ & $0.977$ & $0.984$ & $0.989$ \\
  \hline
\end{tabular}
\end{subtable}

\begin{subtable}[t]{0.5\textwidth}
\caption{Galaxy Nexus-Long short term memory}
\vspace{-0.15cm}
\label{tab:time_pred_lstm2}
\centering
\scriptsize
\begin{tabular}{ |c | c | c | c | c | c | c | } 
 \hline
 & FastDeepIoT & SVR & DT & RF & GBRT & DNN  \\ 
  \hline
   \hline
   MAPE & $\mathbf{2.9\%}$ & $66.8\%$ & $8.4\%$ & $7.8\%$ & $6.0\%$ & $\textbf{2.9\%}$  \\
  \hline
  MAE & $\mathbf{1.4}$ & $26.2$ & $3.0$ & $3.3$ & $2.7$ & $\mathbf{1.3}$ \\
  \hline
  $\text{R}^2$ & $\mathbf{0.997}$ & $-0.196$ & $0.983$ & $0.985$ & $0.987$ & $\mathbf{0.997}$ \\
  \hline
\end{tabular}
\end{subtable}

\begin{subtable}[t]{0.5\textwidth}
\caption{Galaxy Nexus-Fully-connected layer}
\vspace{-0.15cm}
\label{tab:time_pred_fc2}
\centering
\scriptsize
\begin{tabular}{ |c | c | c | c | c | c | c | } 
 \hline
 & FastDeepIoT & SVR & DT & RF & GBRT & DNN  \\ 
  \hline
   \hline
   MAPE & $\mathbf{4.0\%}$ & $55.0\%$ & $12.3\%$ & $11.3\%$ & $9.5\%$ & $4.1\%$  \\
  \hline
  MAE & $\mathbf{0.3}$ & $6.7$ & $1.2$ & $0.9$ & $1.0$ & $\mathbf{0.3}$ \\
  \hline
  $\text{R}^2$ & $\mathbf{0.996}$ & $-0.629$ & $0.944$ & $0.972$ & $0.949$ & $\mathbf{0.996}$ \\
  \hline
\end{tabular}
\end{subtable}
\end{table}
In order to evaluate execution time modeling accuracy, we compare our tree-structured linear regression model to other state-of-the-art regression models on two mobile devices. To evaluate the quality of compression, we present a set of experiments that demonstrate the speed-up of the compressed neural network obtained by the compression steering module with three human-centric interaction and sensing applications.

\vspace{-0.5cm}
\subsection{Execution time Model}~\label{sec:exe_time_exp}
We implement the following execution time estimation alternatives: 
\begin{enumerate}[leftmargin=*,topsep=0pt]
\setlength\itemsep{0pt}
\item \textbf{SVR:} support vector regression with radial basis function kernel~\cite{drucker1997support}. This algorithm tries to perform linear separation over a higher dimensional kernel feature space by characterizing the maximal margin.
\item \textbf{DT:} classification and regression trees~\cite{breiman2017classification}. This is an interpretable model. It groups and predicts execution time by the execution time itself.
\item \textbf{RF:} random forest regression~\cite{breiman2001random}. This algorithm trades the interpretability of regression tree for the predictive performance by ensembling multiple trees with random feature selections.
\item \textbf{GBRT:} gradient boosted regression trees~\cite{friedman2001greedy}. This algorithm builds an additive model in a forward stage-wise fashion, which is hard to interpret.
\item \textbf{DNN:} multilayer perceptron~\cite{lecun2015deep}. Deep neural network is a learning model with high capacity. We build a four-layer fully connected neural network with LeRU as the activation function, except for the output layer. We fine-tune the structure and apply dropout as well as L2 regularization to prevent overfitting. DNN is a black-box model. 
\end{enumerate}

We train all the baseline models with the dataset generated by the profiling module in FastDeepIoT ($75\%$ for training and $25\%$ for testing). For each deep learning component, such as CNN and LSTM, an individual model is trained. We have trained these models with feature vector $\mathbf{f}$, explanatory variable vector $\mathbf{x}$, and the concatenate of feature and explanatory variable vectors as inputs, where $\mathbf{f}$ and $\mathbf{x}$ are the same as the definitions in Section~\ref{sec:build_run_time_model}. We find that the model trained with explanatory variable vector $\mathbf{x}$ outperforms other choices consistently in all cases, so we only report the results of models trained with $\mathbf{x}$ for simplicity.

\begin{table*}[!htb]
\vspace{-0.35cm}
\begin{center}
\caption {VGGNet (hidden units) on CIFAR-10 dataset.}
\vspace{-0.1cm}
\scriptsize
\label{tab:cifar10_vgg}
\vspace{-0.1cm}
\begin{tabular}{ |c | c | c | c | c | c | c | } 
\hline
 & \multicolumn{4}{ c| }{No Execution Time Model} & Nexus 5 & Galaxy Nexus  \\
 \hline
Layer & Original  & DeepIoT & DeepIoT+localMin & DeepIoT+FLOPs & FastDeepIoT & FastDeepIoT \\ 
  \hline
   \hline
   conv1-1 ($3\times3$) & $64$ & $27$ & $28$ & $19$ & $12$ & $16$ \\
  \hline
  conv1-2 ($3\times3$) & $64$ & $47$ & $48$ & $17$ & $16$ & $24$ \\
  \hline
  conv2-1 ($3\times3$) & $128$ & $53$ & $56$ & $33$ & $28$ & $36$  \\
  \hline
  conv2-2 ($3\times3$) & $128$ & $68$ & $68$ & $50$  & $32$ & $44$ \\
  \hline
  conv3-1 ($3\times3$) & $256$ & $104$ & $104$ & $89$  & $64$ & $72$ \\
  \hline
  conv3-2 ($3\times3$) & $256$ & $97$ & $100$ & $79$  & $64$ & $56$ \\
  \hline
  conv3-3 ($3\times3$) & $256$ & $89$ & $92$ & $77$  & $68$ & $72$ \\
  \hline
  conv4-1 ($3\times3$) & $512$ & $122$ & $124$ & $115$  & $132$ & $96$ \\
  \hline
  conv4-2 ($3\times3$) & $512$ & $95$ & $96$ & $112$  & $136$  & $80$ \\
  \hline
  conv4-3 ($3\times3$) & $512$ & $64$ & $64$ & $112$  & $104$ & $120$ \\
  \hline
  conv5-1 ($2\times2$) & $512$ & $128$ & $128$ & $143$  & $148$ & $116$ \\
  \hline
  conv5-2 ($2\times2$) & $512$ & $112$ & $112$ & $132$  & $144$ & $108$ \\
  \hline
  conv5-3 ($2\times2$) & $512$ & $146$ & $148$ & $182$  & $104$ & $92$ \\
  \hline
  fc1 & $4096$ & $27$ & $27$ & $1097$  & $132$ & $132$ \\
  \hline
  fc2 & $4096$ & $161$ & $161$ & $935$  & $152$ & $123$ \\
  \hline
  fc3 & $1000$ & $10$ & $96$ & $72$  & $157$ & $167$ \\
  \hline
  \hline
  Test accuracy & $90.6\%$ & $90.6\%$ & $90.6\%$ & $90.6\%$ & $90.6\%$ & $90.6\%$ \\
  \hline
  Execution time $t$ (Nexus 5) & 328 ms & 31 ms & 21 ms & 28 ms & \textbf{16 ms} &  23 ms  \\
  \hline
  Execution time $t$ (Galaxy Nexus) & 610 ms & 72 ms & 63 ms & 52 ms & 36 ms &  \textbf{34 ms}  \\
  \hline
\end{tabular}
\end{center}
\end{table*} 
\setlength{\textfloatsep}{0pt}

\begin{figure*}[!htb]
\vspace{-0.5cm}
\begin{subfigure}{.32\linewidth}
  \centering
  \includegraphics[width=0.86\linewidth]{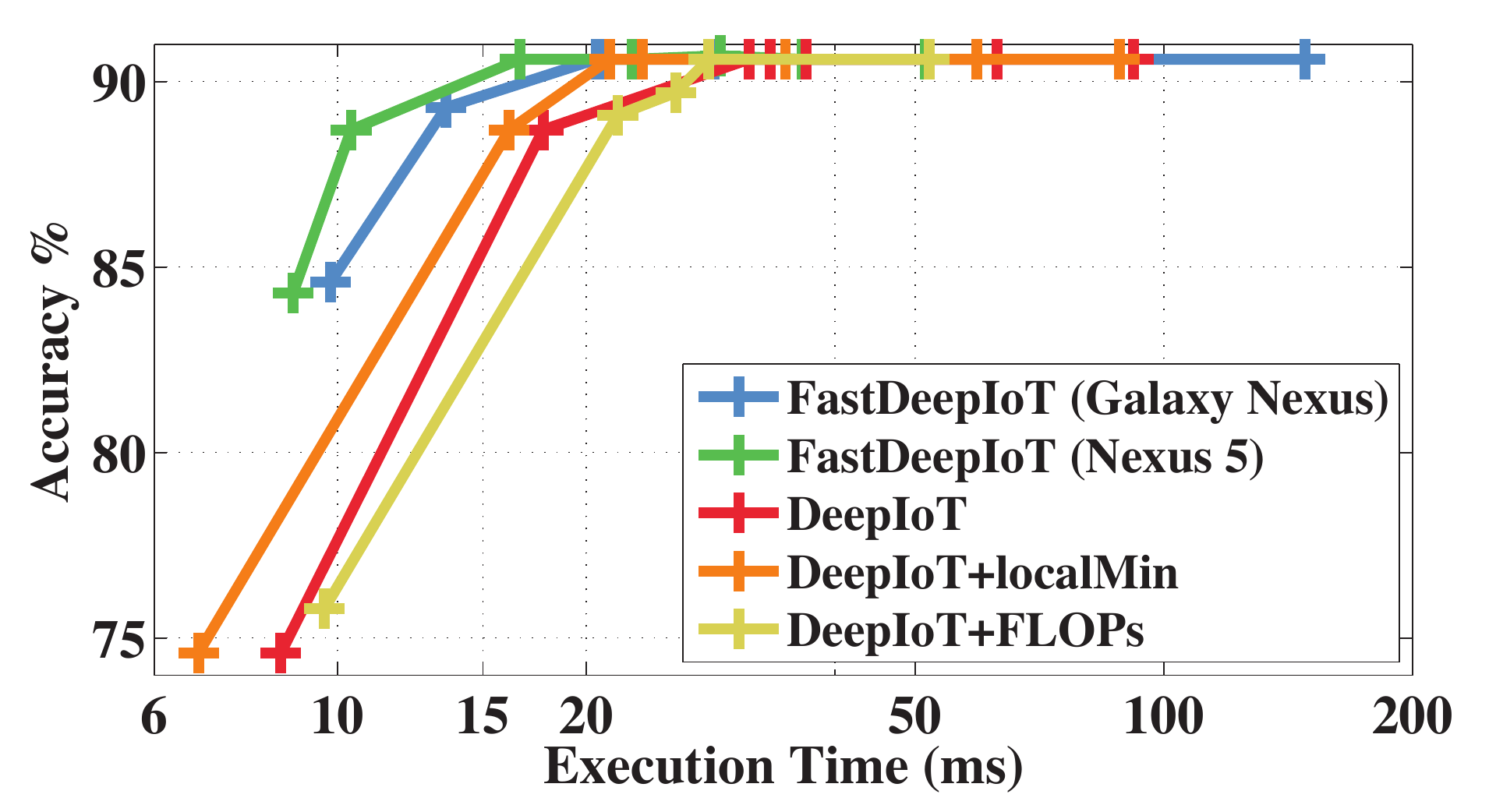}
  \vspace{-0.3cm}
  \caption{Tradeoff between testing accuracy and execution time on Nexus 5. }
  \label{fig:CIFAR10_acc_time_n5}
\end{subfigure}%
\begin{subfigure}{.32\linewidth}
  \centering
  \includegraphics[width=0.86\linewidth]{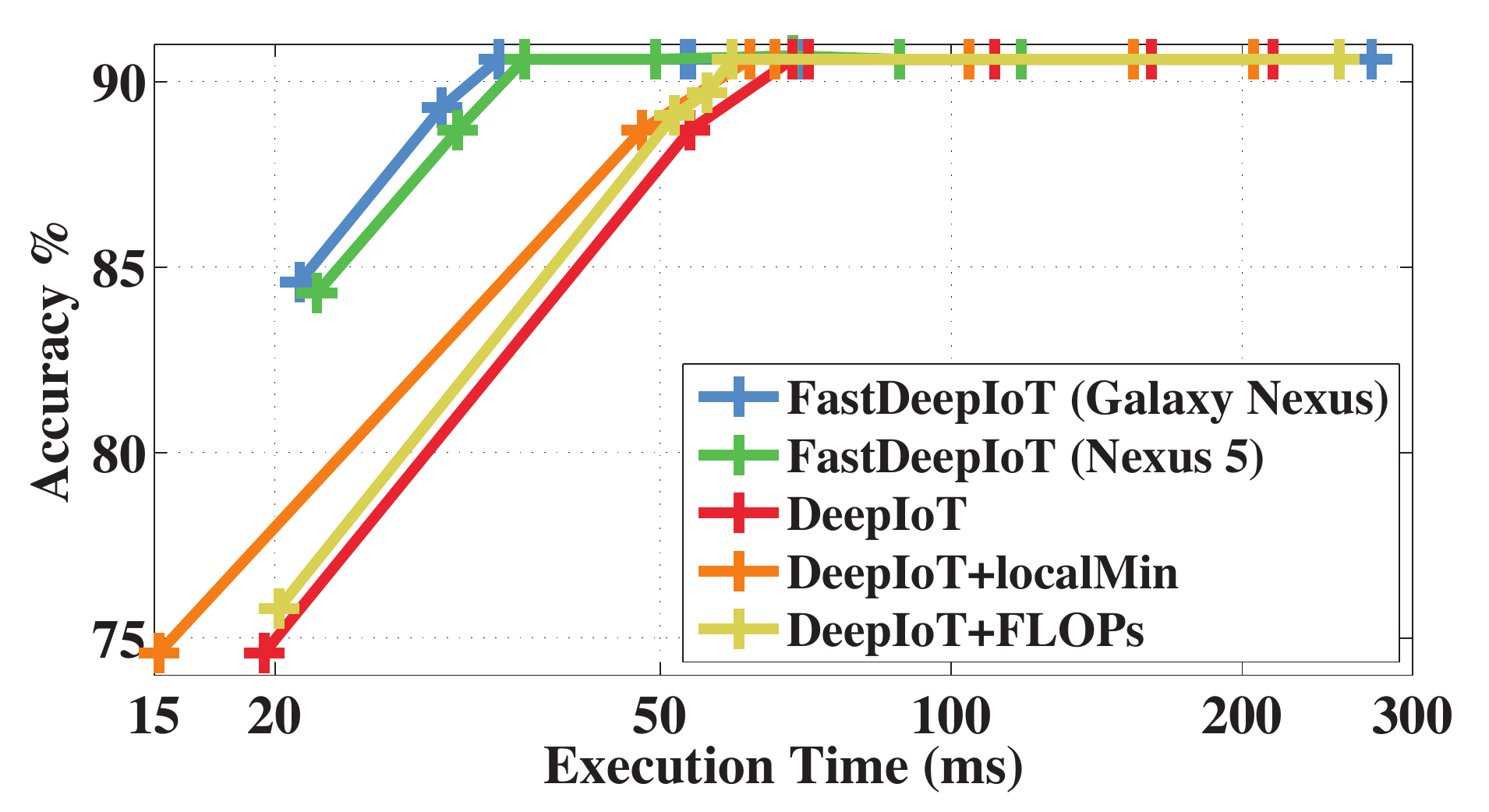}
  \vspace{-0.3cm}
  \caption{Tradeoff between testing accuracy and execution time on Galaxy Nexus.}
  \label{fig:CIFAR10_acc_time_gn}
\end{subfigure}
\begin{subfigure}{.32\linewidth}
  \centering
  \includegraphics[width=0.86\linewidth]{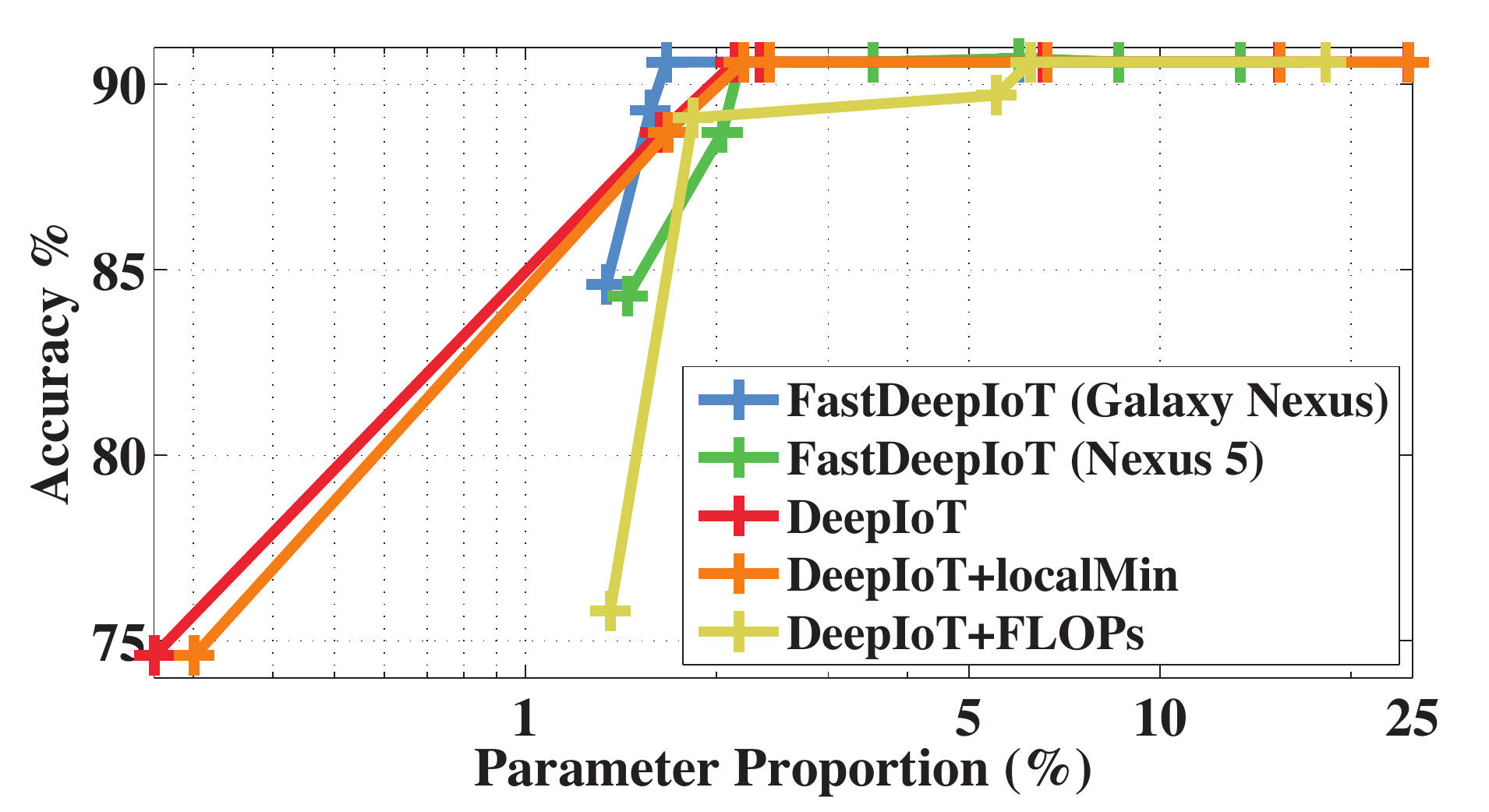}
  \vspace{-0.3cm}
  \caption{Tradeoff between testing accuracy and compressed parameter size.}
  \label{fig:CIFAR10_acc_param}
\end{subfigure}
\begin{subfigure}{.35\linewidth}
  \centering
  \includegraphics[width=0.79\linewidth]{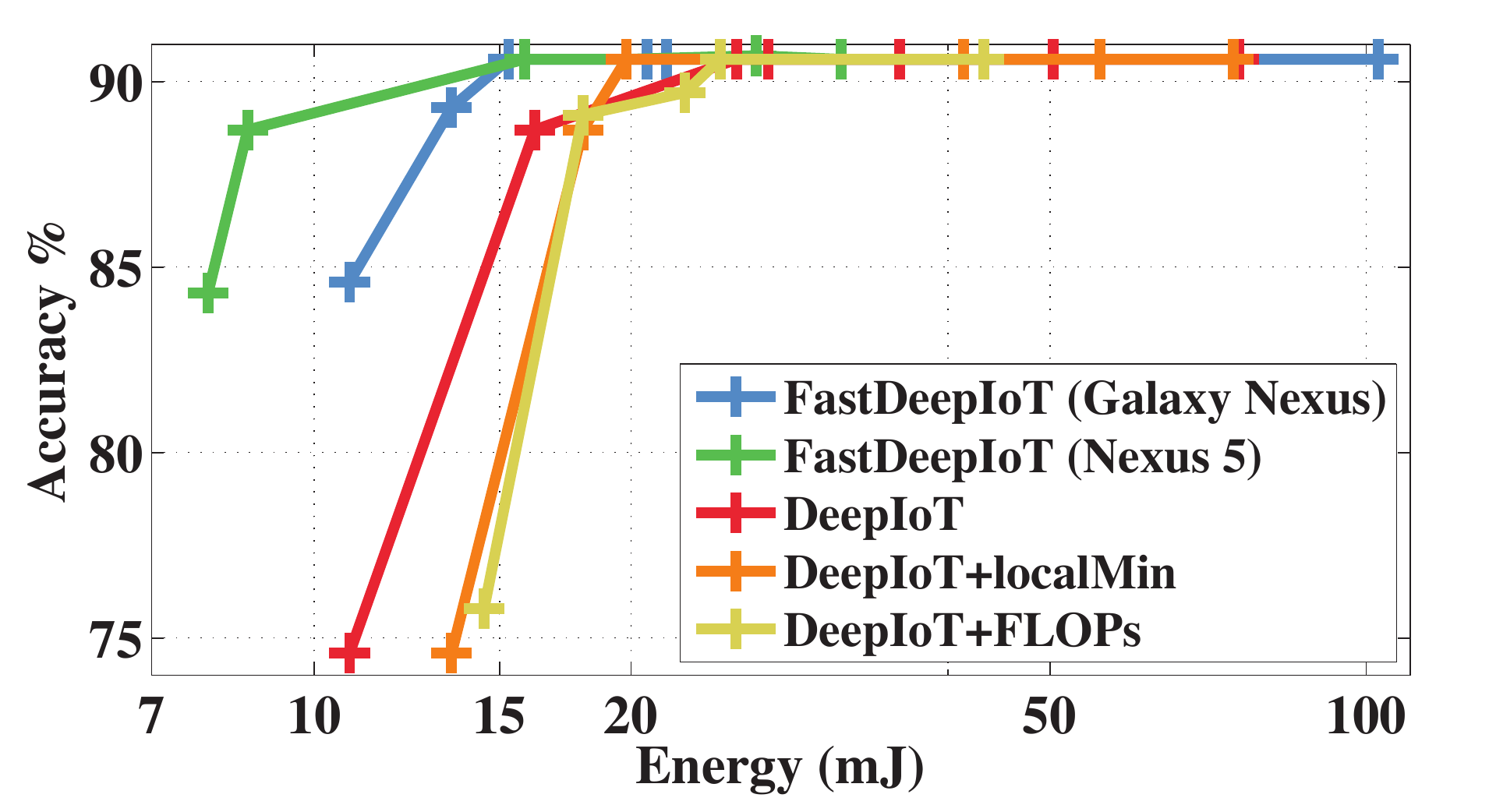}
  \vspace{-0.3cm}
  \caption{Tradeoff between testing accuracy and energy consumption on Nexus 5.}
  \label{fig:CIFAR10_acc_energy_n5}
\end{subfigure}
\begin{subfigure}{.35\linewidth}
  \centering
  \includegraphics[width=0.79\linewidth]{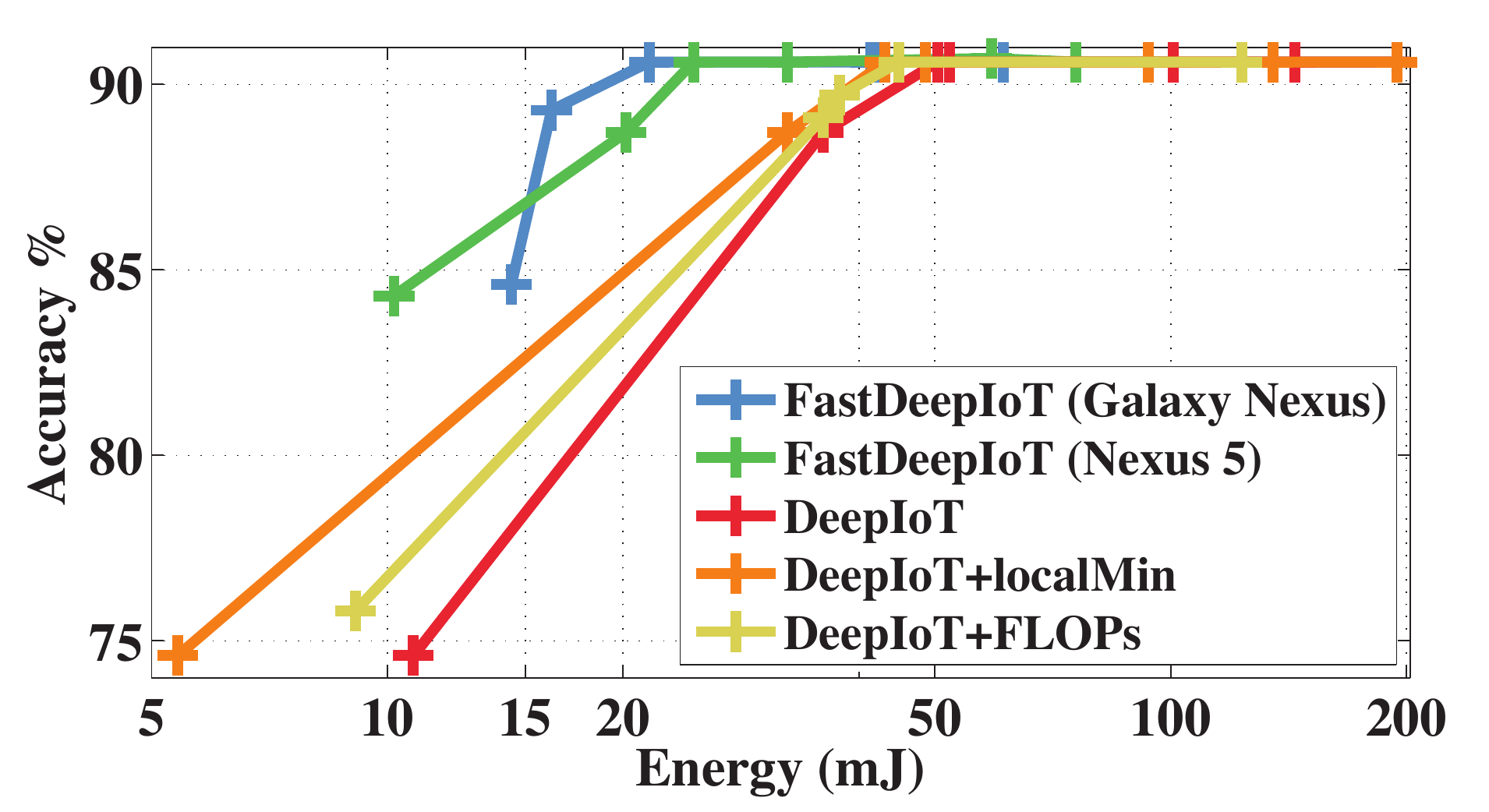}
  \vspace{-0.3cm}
  \caption{Tradeoff between testing accuracy and energy consumption on Galaxy Nexus.}
  \label{fig:CIFAR10_acc_energy_gn}
\end{subfigure}
\vspace{-0.2cm}
\caption{System performance tradeoff for VGGNet on CIFAR-10 dataset}
\label{fig:CIFAR10}
\vspace{-0.5cm}
\end{figure*}

We evaluate these models on convolutional layer, gated recurrent unit, long short term memory, and fully-connected layer with 
mean absolute percentage error, mean absolute rrror, and coefficient of determination on two hardware. As shown in Table~\ref{tab:time_pred}, FastDeepIoT is consistently among top 2 predictors for all experiments with all three metrics. FastDeepIoT also outperforms the highly capable deep learning model for more than half of the cases, while FastDeepIoT is much more interpretable. There are two reasons for the remarkable performance of FastDeepIoT. On one hand, FastDeepIoT captures the primary characters of deep learning execution time behaviours, which makes an interpretable and accurate model possible. 
On the other hand, since the profiled dataset is limited (around one thousand samples for training), complex models such as deep neural networks that require large training dataset may not be the best choice here.

\vspace{-0.3cm}
\subsection{Compression Steering Module}
In this section, we evaluate the performance of our compression steering module with three sensing applications. We train the neural networks on traditional benchmark datasets as original models. Then, we compress the original models using FastDeepIoT and the three state-of-the-art baseline algorithms. Finally, we test the accuracy, execution time, and energy consumption of  compressed models on mobile devices.

We compare FastDeepIoT with three baseline algorithms:
\begin{enumerate}[leftmargin=*,topsep=0pt]
\setlength\itemsep{0pt}
\item \textbf{DeepIoT:} This is a state-of-the-art neural structure compression algorithm~\cite{yao2017deepiot}. The algorithm designs a compressor neural network with adaptive dropout to explore a succinct structure for the original model.
\item \textbf{DeepIoT+localMin:} We enhance DeepIoT with the ability of expanding layer for finding execution-time local minima. This method takes the compressed model of DeepIoT and expands its layers with zero-value elements that can trigger local minima according to Algorithm~\ref{alg:local_minima}. We use this almost zero-effort method to show the improvement made on existing compressed models by interpreting deep learning execution time with FastDeepIoT.
\item \textbf{DeepIoT+FLOPs:} This method enhances DeepIoT by adding a term that minimizes FLOPs to the original objective function~\eqref{eqn:org_compress}. Since a large proportion of works use FLOPs as the execution time estimation~\cite{Zhang2017ShuffleNetAE,Howard2017MobileNetsEC,Iandola2016SqueezeNetAA}, this method shows to what extend FLOPs can be used to compress neural network for reducing execution time.
\end{enumerate}

\vspace{-0.3cm}
\subsubsection{Image recognition on CIFAR-10}
This is a vision based task, image recognition based on a low-resolution camera. During this experiment, we use CIFAR-10 as our training and testing dataset. The CIFAR-10 dataset consists of 60000 $32\times32$ colour images in 10 classes, with 6000 images per class. There are 50000 training images and 10000 test images. 

\begin{table*}[!htb]
\vspace{-0.4cm}
\begin{center}
\caption {VGGNet (hidden units) on ImageNet dataset.}
\vspace{-0.15cm}
\scriptsize
\label{tab:imagenet_vgg}
\vspace{-0.1cm}
\begin{tabular}{ |c | c | c | c | c | c | c | } 
\hline
 & \multicolumn{4}{ c| }{No Execution Time Model} & Nexus 5 & Galaxy Nexus  \\
 \hline
Layer & Original  & DeepIoT & DeepIoT+localMin & DeepIoT+FLOPs & FastDeepIoT & FastDeepIoT \\ 
  \hline
   \hline
   conv1-1 ($3\times3$) & $64$ & 43 & 44 & 23 & 12 & 16 \\
  \hline
  conv1-2 ($3\times3$) & $64$ & 47 & 48 & 32 & 12 & 16 \\
  \hline
  conv2-1 ($3\times3$) & $128$ & 100 & 100 & 65 & 20 & 44  \\
  \hline
  conv2-2 ($3\times3$) & $128$ & 97 & 100 & 67 & 40 & 40 \\
  \hline
  conv3-1 ($3\times3$) & $256$ & 164 & 164 & 116 & 88 & 108 \\
  \hline
  conv3-2 ($3\times3$) & $256$ & 164 & 164 & 135 & 72 & 104 \\
  \hline
  conv3-3 ($3\times3$) & $256$ & 153 & 156 & 70 & 116 & 108 \\
  \hline
  conv4-1 ($3\times3$) & $512$ & 235 & 236 & 72 & 268 & 240 \\
  \hline
  conv4-2 ($3\times3$) & $512$ & 240 & 240 & 181 & 236 & 216 \\
  \hline
  conv4-3 ($3\times3$) & $512$ & 220 & 240 & 258 & 340 & 200 \\
  \hline
  conv5-1 ($3\times3$) & $512$ & 255 & 256 & 261 & 376 & 240 \\
  \hline
  conv5-2 ($3\times3$) & $512$ & 260 & 260 &303  & 376 & 288 \\
  \hline
  conv5-3 ($3\times3$) & $512$ & 257 & 260 & 47 & 176 & 216 \\
  \hline
  fc1 & $4096$ & 436 & 436 & 1594 & 656 & 920 \\
  \hline
  fc2 & $4096$ & 1169 & 1169 & 824 & 1150 & 1189 \\
  \hline
  fc3 & $1000$ & 297 & 297 & 405 & 287 & 402 \\
  \hline
  \hline
  Test top-5 accuracy & $88.9\%$  & $88.9\%$ & $88.9\%$ & $88.9\%$ & $88.9\%$ & $88.9\%$ \\
  \hline
  Execution time $t$ (Nexus 5)  & \diagbox[dir=SW,width=1.15cm, height=0.25cm]{}{} & 1682 ms & 1605 ms & 968.8 ms & \textbf{688.8 ms} & 725.7 ms \\
  \hline
  Execution time $t$ (Galaxy Nexus) & \diagbox[dir=SW,width=1.15cm, height=0.25cm]{}{} & 7773 ms & 6991 ms & 3930 ms & 3211 ms & \textbf{2930 ms} \\
  \hline
\end{tabular}
\end{center}
\end{table*} 
\setlength{\textfloatsep}{0pt}

\begin{figure*}[!htb]
\vspace{-0.5cm}
\begin{subfigure}{.32\linewidth}
  \centering
  \includegraphics[width=0.86\linewidth]{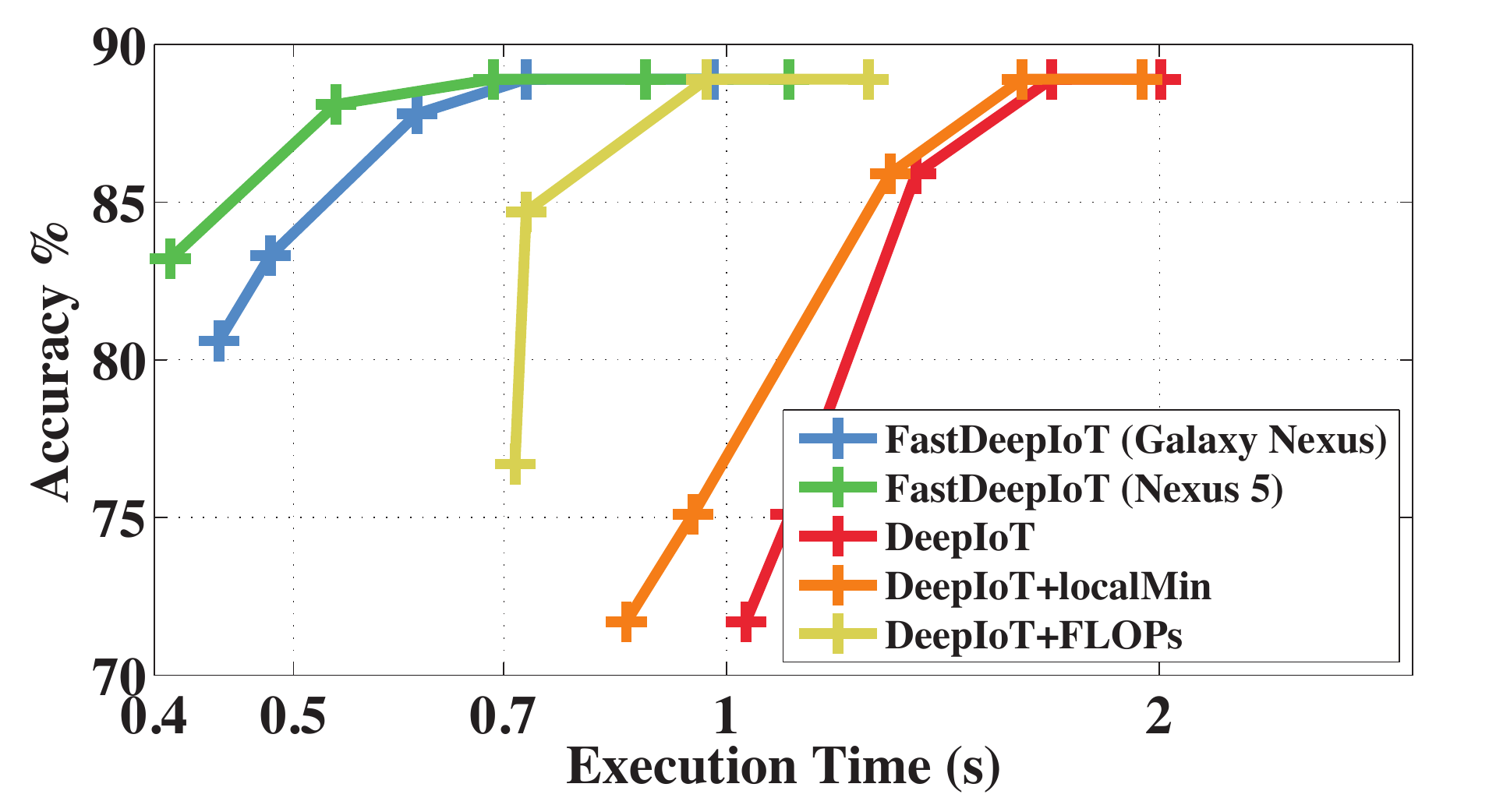}
  \vspace{-0.3cm}
  \caption{Tradeoff between testing accuracy and execution time on Nexus 5. }
  \label{fig:imageNet_acc_time_n5}
\end{subfigure}%
\begin{subfigure}{.32\linewidth}
  \centering
  \includegraphics[width=0.86\linewidth]{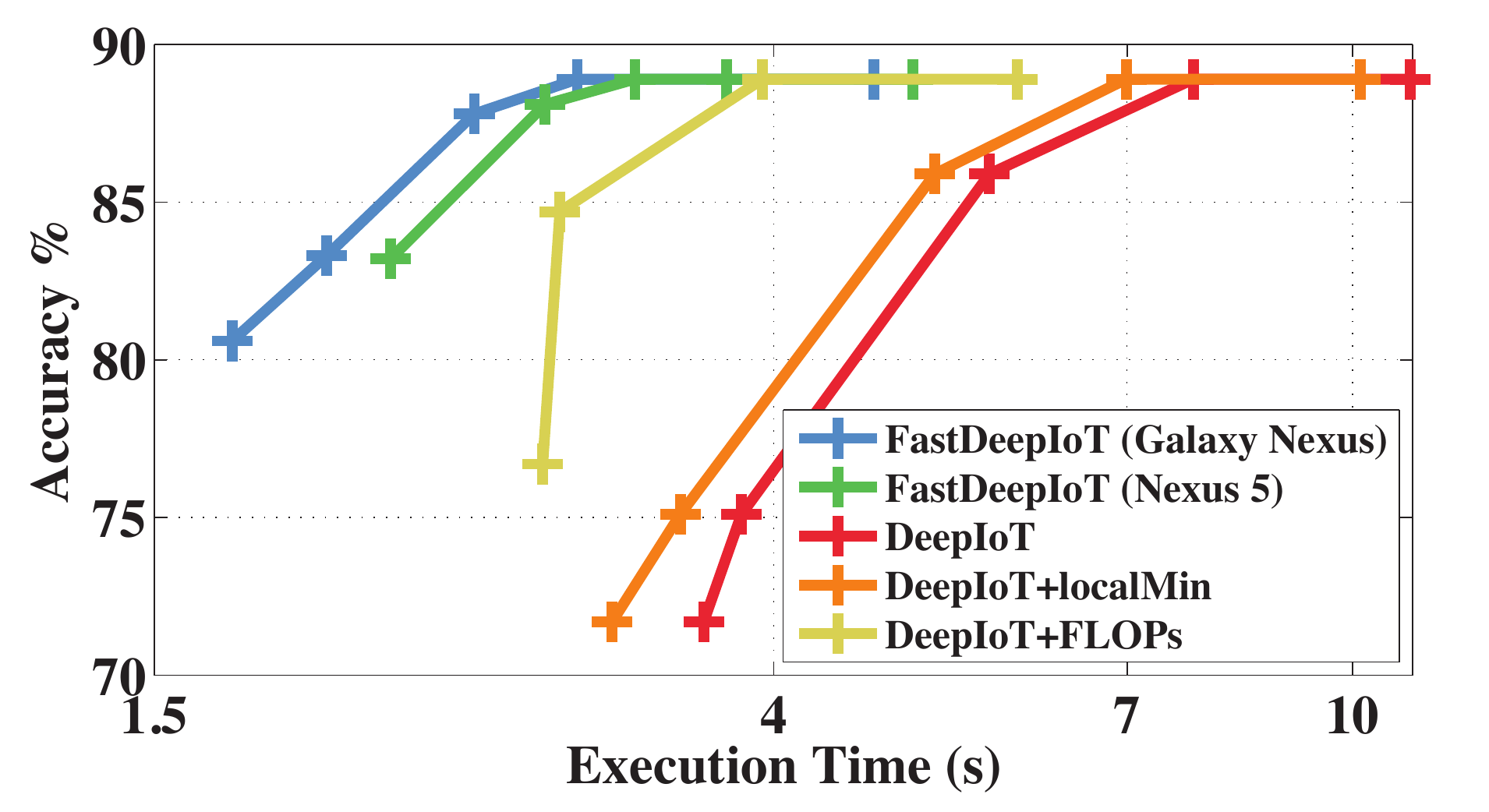}
  \vspace{-0.3cm}
  \caption{Tradeoff between testing accuracy and execution time on Galaxy Nexus.}
  \label{fig:imageNet_acc_time_gn}
\end{subfigure}
\begin{subfigure}{.32\linewidth}
  \centering
  \includegraphics[width=0.86\linewidth]{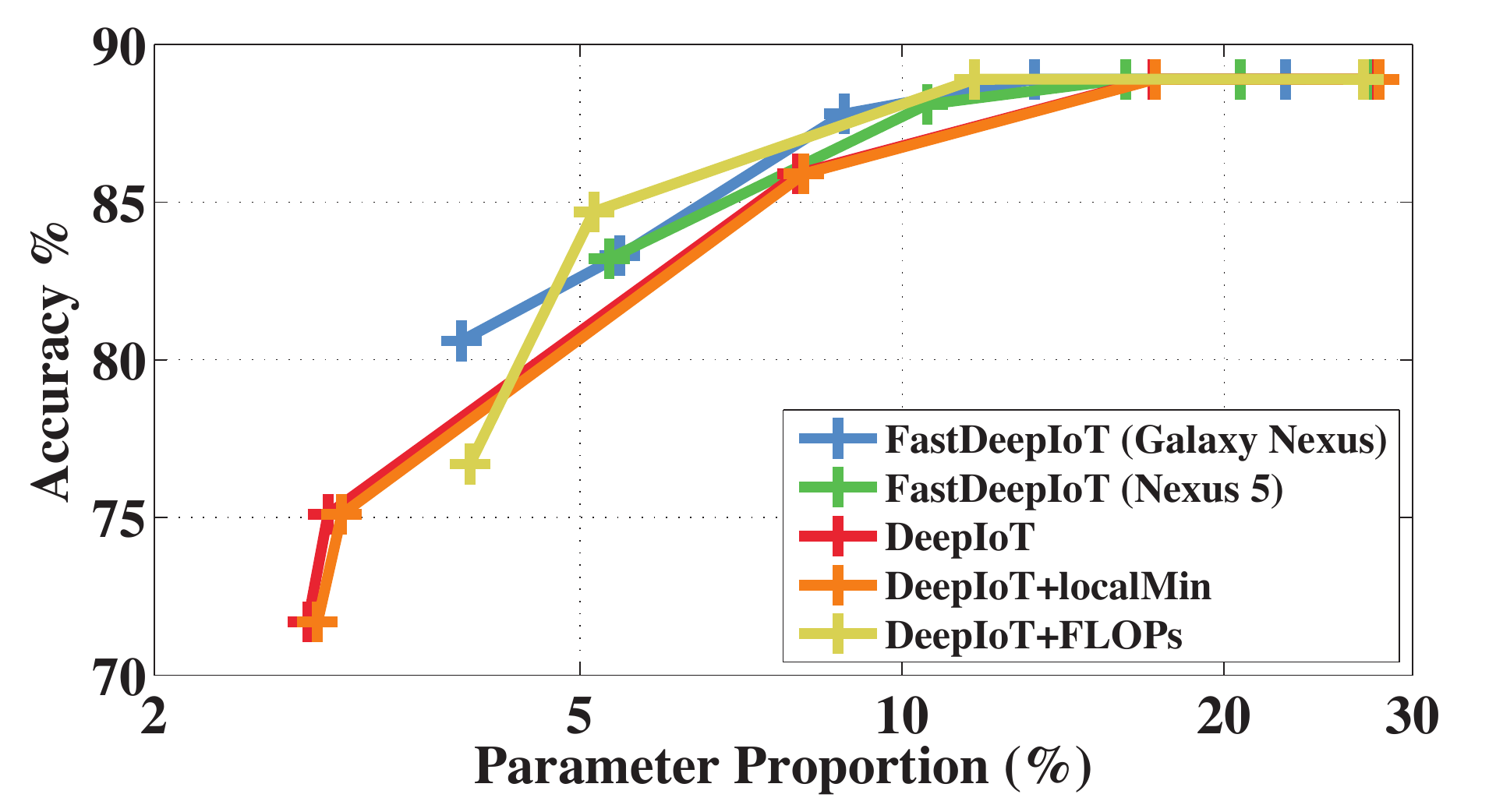}
  \vspace{-0.3cm}
  \caption{Tradeoff between testing accuracy and compressed parameter size.}
  \label{fig:imageNet_acc_param}
\end{subfigure}
\begin{subfigure}{.35\linewidth}
  \centering
  \includegraphics[width=0.79\linewidth]{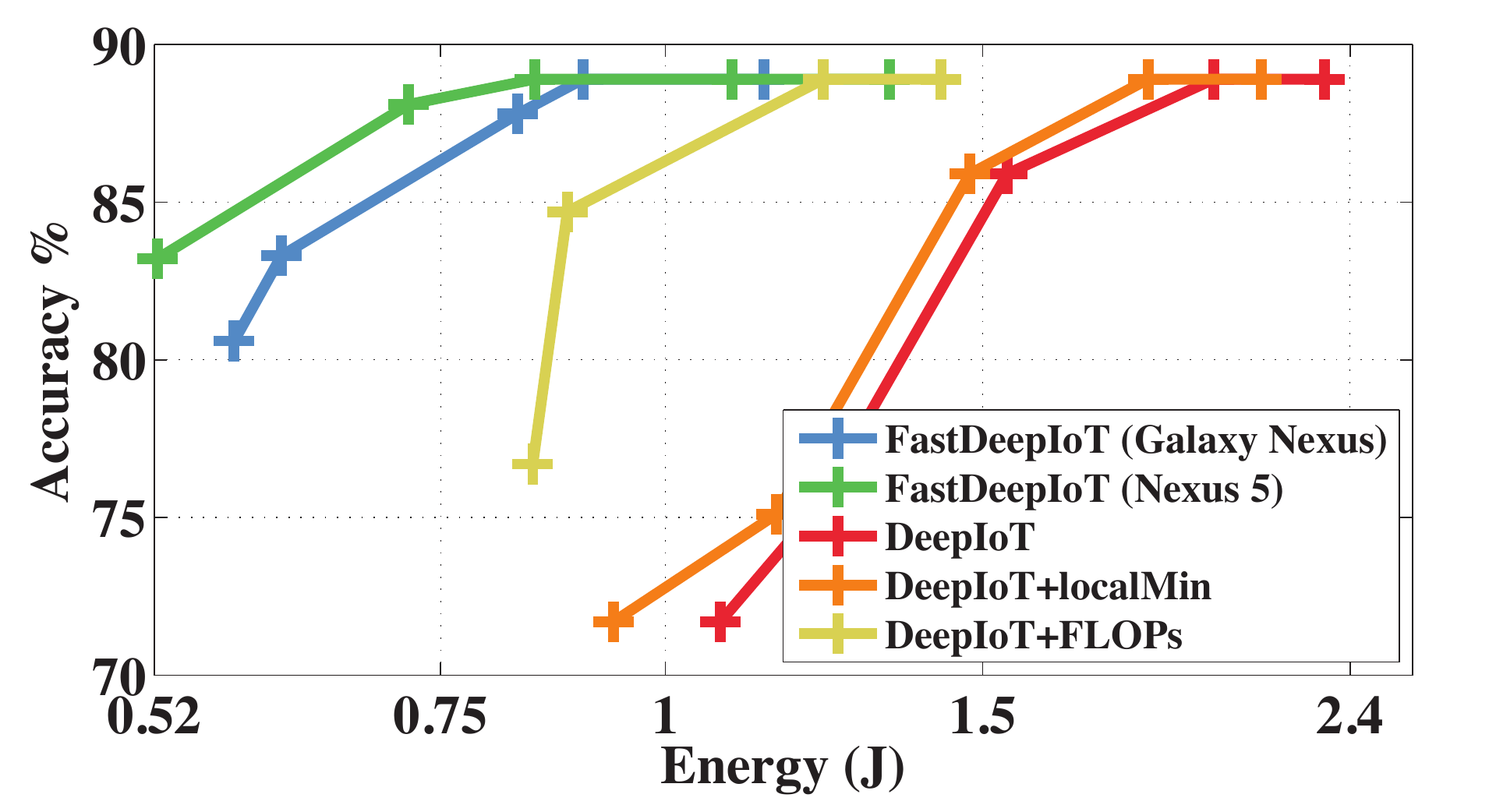}
  \vspace{-0.3cm}
  \caption{Tradeoff between testing accuracy and energy consumption on Nexus 5.}
  \label{fig:imageNet_acc_energy_n5}
\end{subfigure}
\begin{subfigure}{.35\linewidth}
  \centering
  \includegraphics[width=0.79\linewidth]{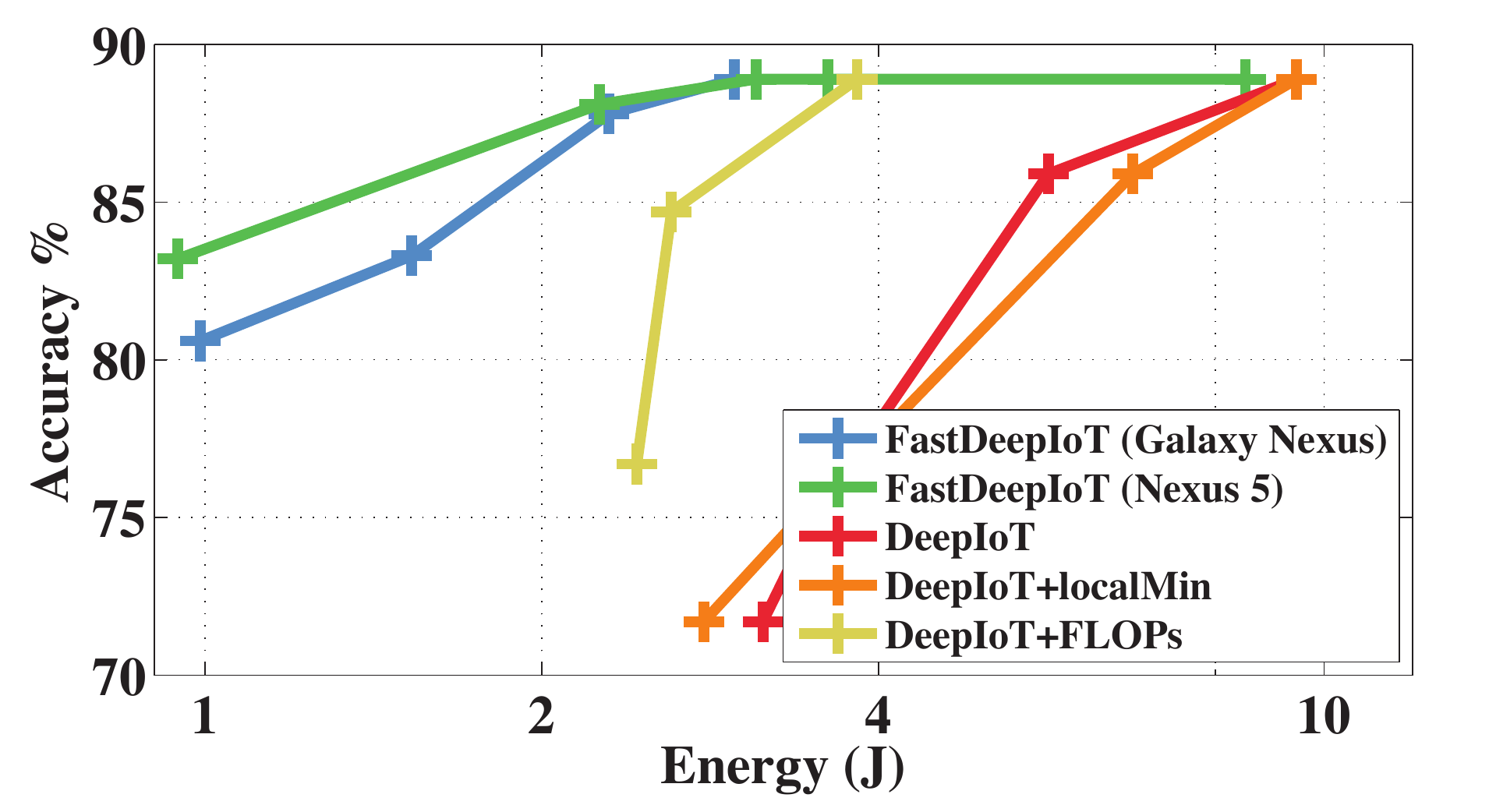}
  \vspace{-0.3cm}
  \caption{Tradeoff between testing accuracy and energy consumption on Galaxy Nexus.}
  \label{fig:imageNet_acc_energy_gn}
\end{subfigure}
\vspace{-0.2cm}
\caption{System performance tradeoff for VGGNet on ImageNet dataset}
\vspace{-0.4cm}
\label{fig:imageNet}
\end{figure*}

During the evaluation, we use VGGNet structure as the original network structure~\cite{simonyan2014very}. The detailed structure is shown in Table~\ref{tab:cifar10_vgg}, where we also illustrate the best compressed models that keeps the original test accuracy for all algorithms. The compressed model can be even deployed on tiny IoT devices such as Intel Edison.

As shown in Table~\ref{tab:cifar10_vgg}, FastDeepIoT achieves the best performance on two hardware with their corresponding execution time models. Compared with the state-of-the-art DeepIoT algorithm, FastDeepIoT can further reduce the model execution time by $48\%$ to $53\%$. DeepIoT+localMin outperforms DeepIoT on two hardware, reducing the execution time by $12\%$ to $32\%$. This shows that we can decently reduce the neural network execution time by simply expanding the neural network structure to local execution-time minima. In additional, DeepIoT+FLOPs can speed up the model execution time compared with DeepIoT. However, FastDeepIoT still outperforms DeepIoT+FLOPs by a significant margin. This result highlights that FLOPs is not a proper estimation of time.

Figure~\ref{fig:CIFAR10_acc_time_n5} and~\ref{fig:CIFAR10_acc_time_gn}  shows the tradeoff between testing  accuracy and execution time for different algorithms. FastDeepIoT consistently outperforms other algorithms by a significant margin. Furthermore, the execution time characters on different hardware can affect the final performance. FastDeepIoT (Nexus 5/Galaxy Nexus) performs better on its corresponding hardware. DeepIoT+localMin achieves a better tradeoff compared with DeepIoT. Therefore, utilizing execution-time local minima is a low-cost strategy to speed up neural network execution. In addition, since FLOPs has different degrees of execution time contribution on different hardware, DeepIoT+FLOPs are not able to achieve a better tradeoff than DeepIoT on all devices.

Figure~\ref{fig:CIFAR10_acc_energy_n5} and~\ref{fig:CIFAR10_acc_energy_gn}  shows the tradeoff between testing accuracy and energy consumption for different algorithms. Although FastDeepIoT is not designed to minimize the energy consumption, FastDeepIoT still achieves the best tradeoff. However, we can see that the characters of energy consumption of deep neural network are different from the execution time. FastDeepIoT with the hardware-specific time models are not always the most energy-saving method on the corresponding hardware.  Execution-time local minima cannot consistently help DeepIoT+localMin to outperform DeepIoT. Therefore, further studies on understanding and minimizing deep learning energy consumption are needed.

\begin{table*}[!htb]
\vspace{-0.45cm}
\begin{center}
\caption {DeepSense (hidden units) on HHAR dataset.}
\vspace{-0.15cm}
\scriptsize
\label{tab:HHAR}
\vspace{-0.1cm}
\begin{tabular}{ |c | c | c | c | c | c | c | c | c | c | c | c | c | c | c | c | } 
\hline
 \multicolumn{2}{ |c| }{} & \multicolumn{10}{ c| }{No Execution Time Model} & \multicolumn{2}{ c| }{Nexus 5} & \multicolumn{2}{ c| }{Galaxy Nexus}  \\
 \hline
 \multicolumn{2}{ |c| }{Layer} & \multicolumn{2}{ c| }{Original}  & \multicolumn{2}{ c| }{DeepIoT} & \multicolumn{2}{ c| }{DeepIoT+localMin} & \multicolumn{2}{ c| }{DeepIoT+FLOPs} & \multicolumn{2}{ c| }{\phantom{a}$t_{\text{min}}$\phantom{a}} & \multicolumn{2}{ c| }{FastDeepIoT} & \multicolumn{2}{ c| }{FastDeepIoT} \\ 
  \hline
   \hline
   conv1a & conv1b ($2\times9$) & 64 & 64 & 20 & 19 & $\phantom{11}$20$\phantom{11}$ & 20 & $\phantom{\tiny{1_1}}$26$\phantom{\tiny{1_1}}$ & 25 & $\phantom{\tiny{1_1}}$4$\phantom{\tiny{1_1}}$ & $\phantom{1}$4$\phantom{1}$ & $\phantom{\tiny{1}}$16$\phantom{\tiny{1}}$ & 8 & $\phantom{\tiny{1}}$16$\phantom{\tiny{1}}$  & $\phantom{\tiny{1}}$16$\phantom{\tiny{1}}$  \\
   \hline
   conv2a & conv2b ($1\times3$) & 64 & 64 & 20 & 14 & 20 & 16 & 19 & 17 & 4 & 4 & 8 & 12 & 20 & 16  \\
   \hline
   conv3a & conv3b ($1\times3$) & 64 & 64 & 23 & 23 & 24 & 24 & 22 & 22 & 4 & 4 & 16 &16 & 16 & 16  \\
   \hline
   \multicolumn{2}{ |c| }{conv4 ($2\times8$)} & \multicolumn{2}{ c| }{64} & \multicolumn{2}{ c| }{10} & \multicolumn{2}{ c| }{12} & \multicolumn{2}{ c| }{9} & \multicolumn{2}{ c| }{4} & \multicolumn{2}{ c| }{12} & \multicolumn{2}{ c| }{16}  \\
   \hline
   \multicolumn{2}{ |c| }{conv5 ($1\times6$)} & \multicolumn{2}{ c| }{64} & \multicolumn{2}{ c| }{12} & \multicolumn{2}{ c| }{12} & \multicolumn{2}{ c| }{13} & \multicolumn{2}{ c| }{4} & \multicolumn{2}{ c| }{16} & \multicolumn{2}{ c| }{16}  \\
   \hline
    \multicolumn{2}{ |c| }{conv6 ($1\times4$)} & \multicolumn{2}{ c| }{64} & \multicolumn{2}{ c| }{17} & \multicolumn{2}{ c| }{18} & \multicolumn{2}{ c| }{18} & \multicolumn{2}{ c| }{4} & \multicolumn{2}{ c| }{12} & \multicolumn{2}{ c| }{16}  \\
   \hline
    \multicolumn{2}{ |c| }{gru1} & \multicolumn{2}{ c| }{120} & \multicolumn{2}{ c| }{27} & \multicolumn{2}{ c| }{27} & \multicolumn{2}{ c| }{11} & \multicolumn{2}{ c| }{1} & \multicolumn{2}{ c| }{15} & \multicolumn{2}{ c| }{10} \\
   \hline
   \multicolumn{2}{ |c| }{gru2} & \multicolumn{2}{ c| }{120} & \multicolumn{2}{ c| }{31} & \multicolumn{2}{ c| }{31} & \multicolumn{2}{ c| }{15} & \multicolumn{2}{ c| }{1} & \multicolumn{2}{ c| }{17} & \multicolumn{2}{ c| }{10} \\
   \hline
   \hline
   \multicolumn{2}{ |c| }{Test accuracy} & \multicolumn{2}{ c| }{$94.6\%$} & \multicolumn{2}{ c| }{$94.7\%$} &  \multicolumn{2}{ c| }{$94.7\%$} &  \multicolumn{2}{ c| }{$94.7\%$} &  \multicolumn{2}{ c| }{$16.7\%$} &  \multicolumn{2}{ c| }{$94.7\%$} &  \multicolumn{2}{ c| }{$94.7\%$} \\
   \hline
   \multicolumn{2}{ |c| }{Execution time $t$ (Nexus 5)} & \multicolumn{2}{ c| }{26.2 ms} & \multicolumn{2}{ c| }{19.5 ms} & \multicolumn{2}{ c| }{17.9 ms} & \multicolumn{2}{ c| }{18.3 ms} & \multicolumn{2}{ c| }{14.1 ms} & \multicolumn{2}{ c| }{\textbf{15.3 ms}} & \multicolumn{2}{ c| }{15.8 ms} \\
   \hline
   \multicolumn{2}{ |c| }{$t-t_{\text{min}}$ (Nexus 5)} & \multicolumn{2}{ c| }{12.1 ms} & \multicolumn{2}{ c| }{5.4 ms} & \multicolumn{2}{ c| }{3.8 ms} & \multicolumn{2}{ c| }{4.2 ms} & \multicolumn{2}{ c| }{\diagbox[dir=SW,width=1.75cm, height=0.25cm]{}{}} & \multicolumn{2}{ c| }{\textbf{1.2 ms}} & \multicolumn{2}{ c| }{1.7 ms} \\
   \hline
   \multicolumn{2}{ |c| }{Execution time $t$ (Galaxy Nexus)} & \multicolumn{2}{ c| }{70.9 ms} & \multicolumn{2}{ c| }{30.1 ms} & \multicolumn{2}{ c| }{27.4 ms} & \multicolumn{2}{ c| }{28.2 ms} & \multicolumn{2}{ c| }{18.4 ms} & \multicolumn{2}{ c| }{22.6 ms} & \multicolumn{2}{ c| }{\textbf{22.0 ms}} \\
   \hline
   \multicolumn{2}{ |c| }{$t-t_{\text{min}}$ (Galaxy Nexus)} & \multicolumn{2}{ c| }{52.5 ms} & \multicolumn{2}{ c| }{11.7 ms} & \multicolumn{2}{ c| }{9.0 ms} & \multicolumn{2}{ c| }{9.8 ms} & \multicolumn{2}{ c| }{ \diagbox[dir=SW,width=1.75cm, height=0.25cm]{}{}} & \multicolumn{2}{ c| }{4.2 ms} & \multicolumn{2}{ c| }{\textbf{3.6 ms}} \\
   \hline
\end{tabular}
\end{center}
\end{table*} 
\setlength{\textfloatsep}{0pt}

\begin{figure*}[!htb]
\vspace{-0.5cm}
\begin{subfigure}{.32\linewidth}
  \centering
  \includegraphics[width=0.86\linewidth]{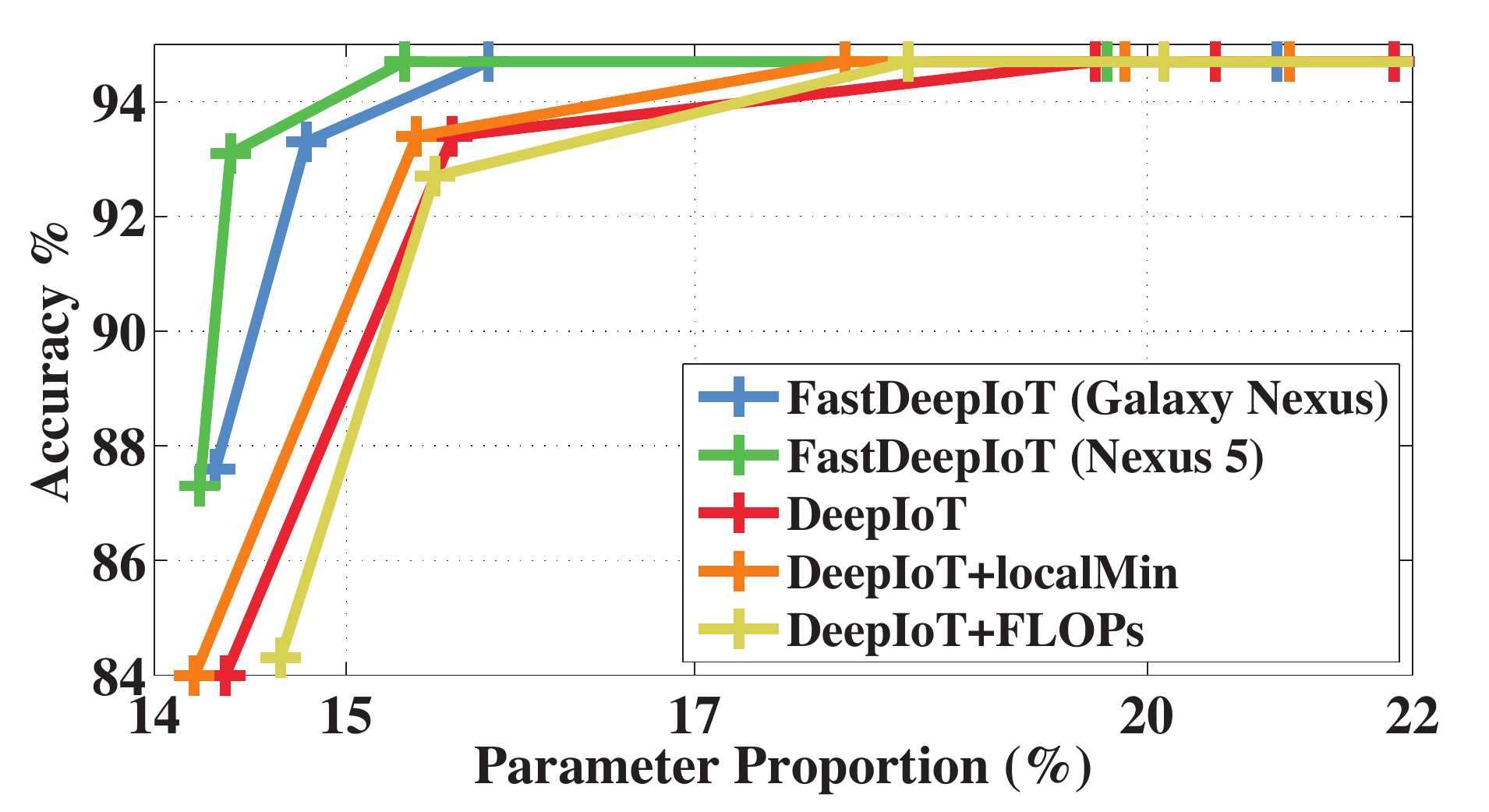}
  \vspace{-0.3cm}
  \caption{Tradeoff between testing accuracy and execution time on Nexus 5. }
  \label{fig:HHAR_acc_time_n5}
\end{subfigure}%
\begin{subfigure}{.32\linewidth}
  \centering
  \includegraphics[width=0.86\linewidth]{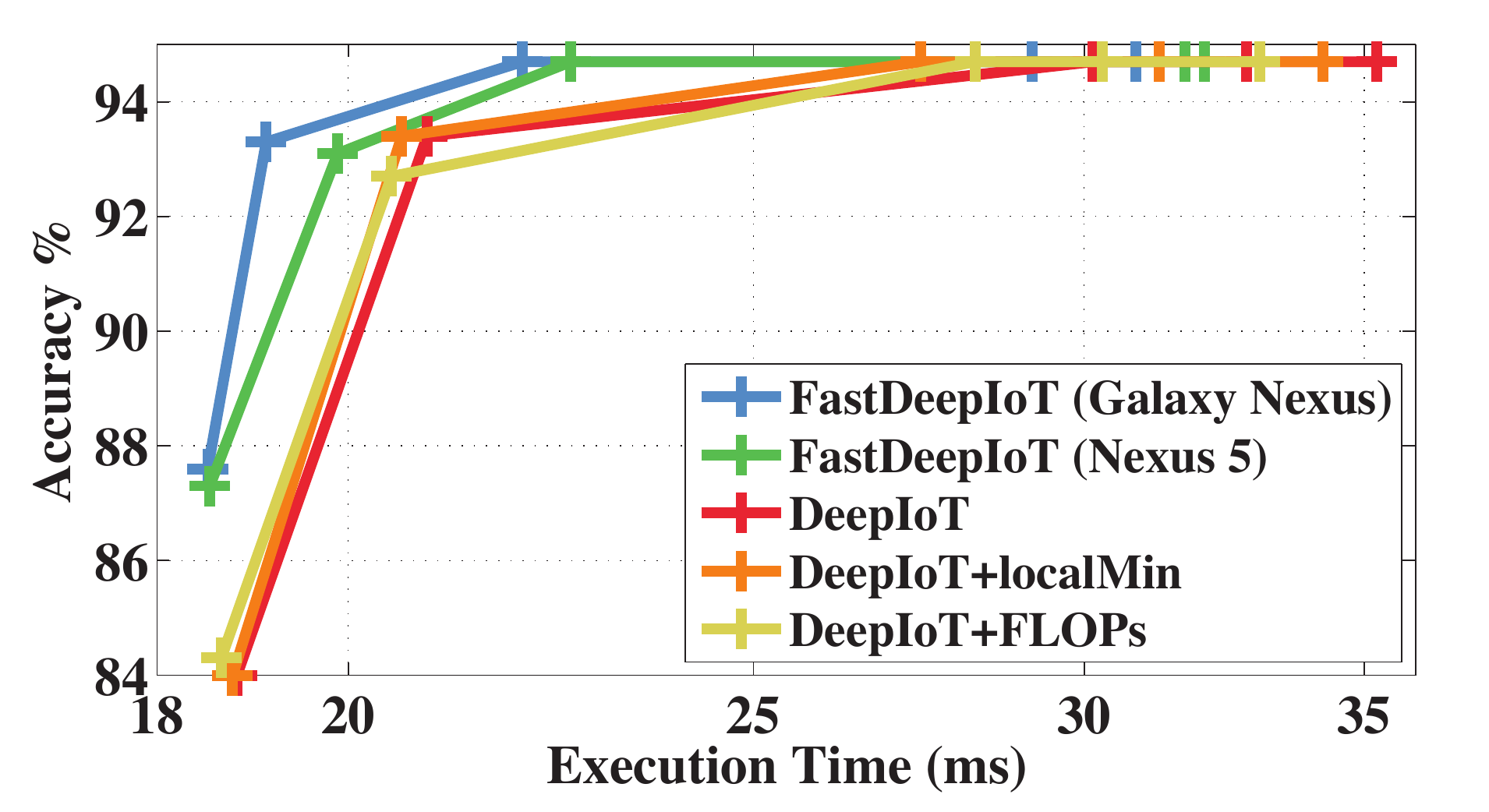}
  \vspace{-0.3cm}
  \caption{Tradeoff between testing accuracy and execution time on Galaxy Nexus.}
  \label{fig:HHAR_acc_time_gn}
\end{subfigure}
\begin{subfigure}{.32\linewidth}
  \centering
  \includegraphics[width=0.86\linewidth]{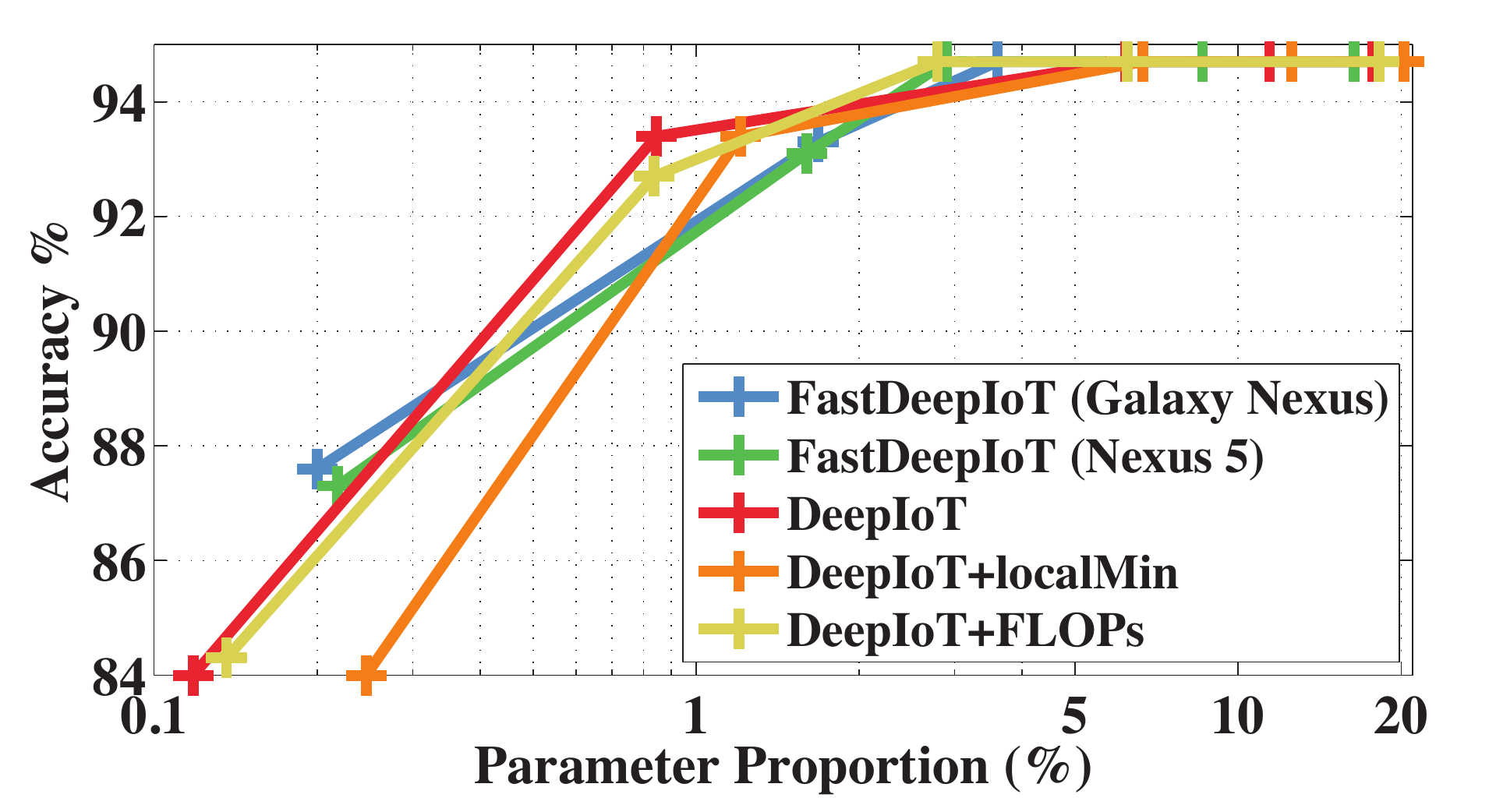}
  \vspace{-0.3cm}
  \caption{Tradeoff between testing accuracy and compressed parameter size.}
  \label{fig:HHAR_acc_param}
\end{subfigure}
\begin{subfigure}{.35\linewidth}
  \centering
  \includegraphics[width=0.79\linewidth]{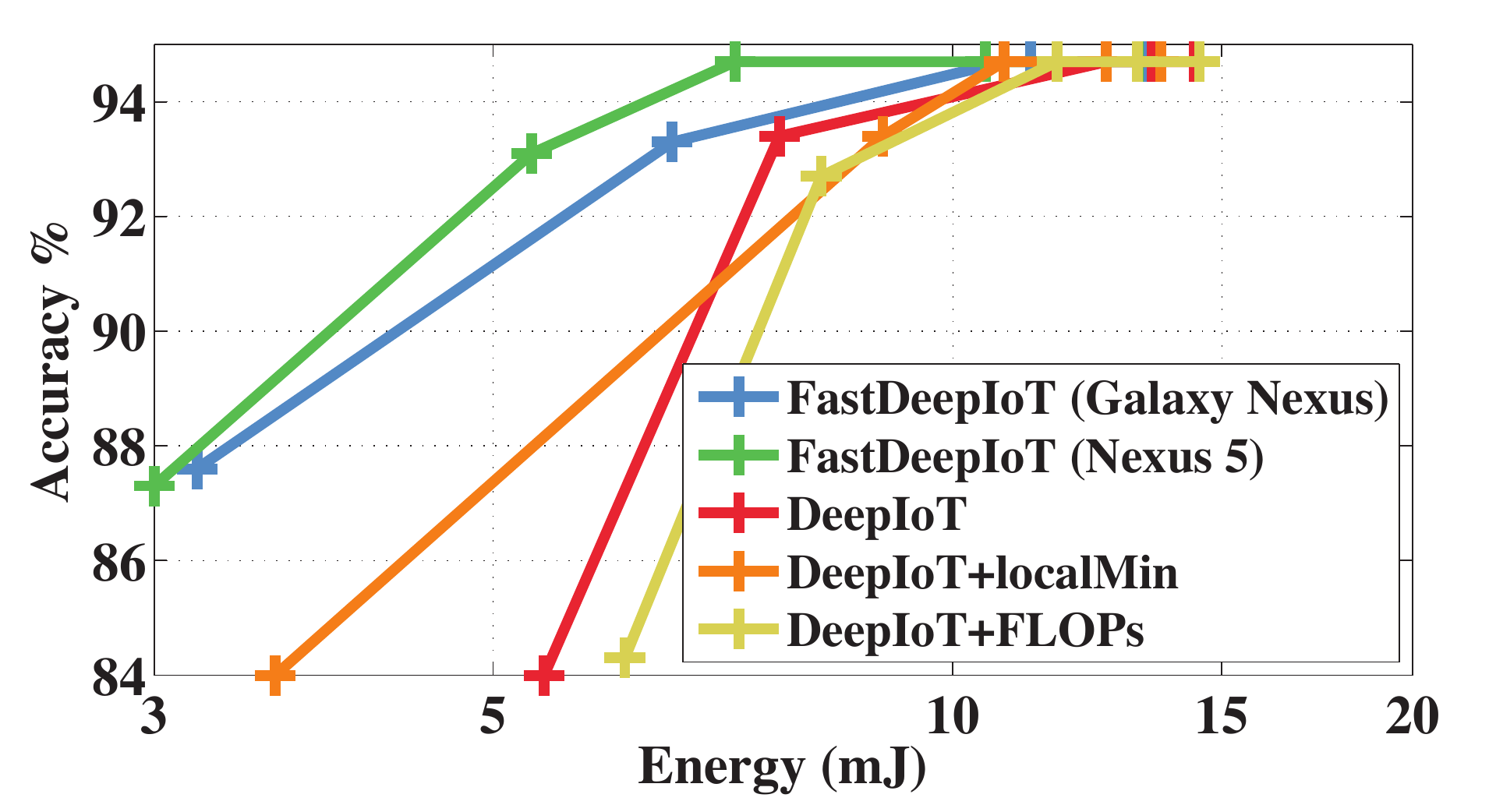}
  \vspace{-0.3cm}
  \caption{Tradeoff between testing accuracy and energy consumption on Nexus 5.}
  \label{fig:HHAR_acc_energy_n5}
\end{subfigure}
\begin{subfigure}{.35\linewidth}
  \centering
  \includegraphics[width=0.79\linewidth]{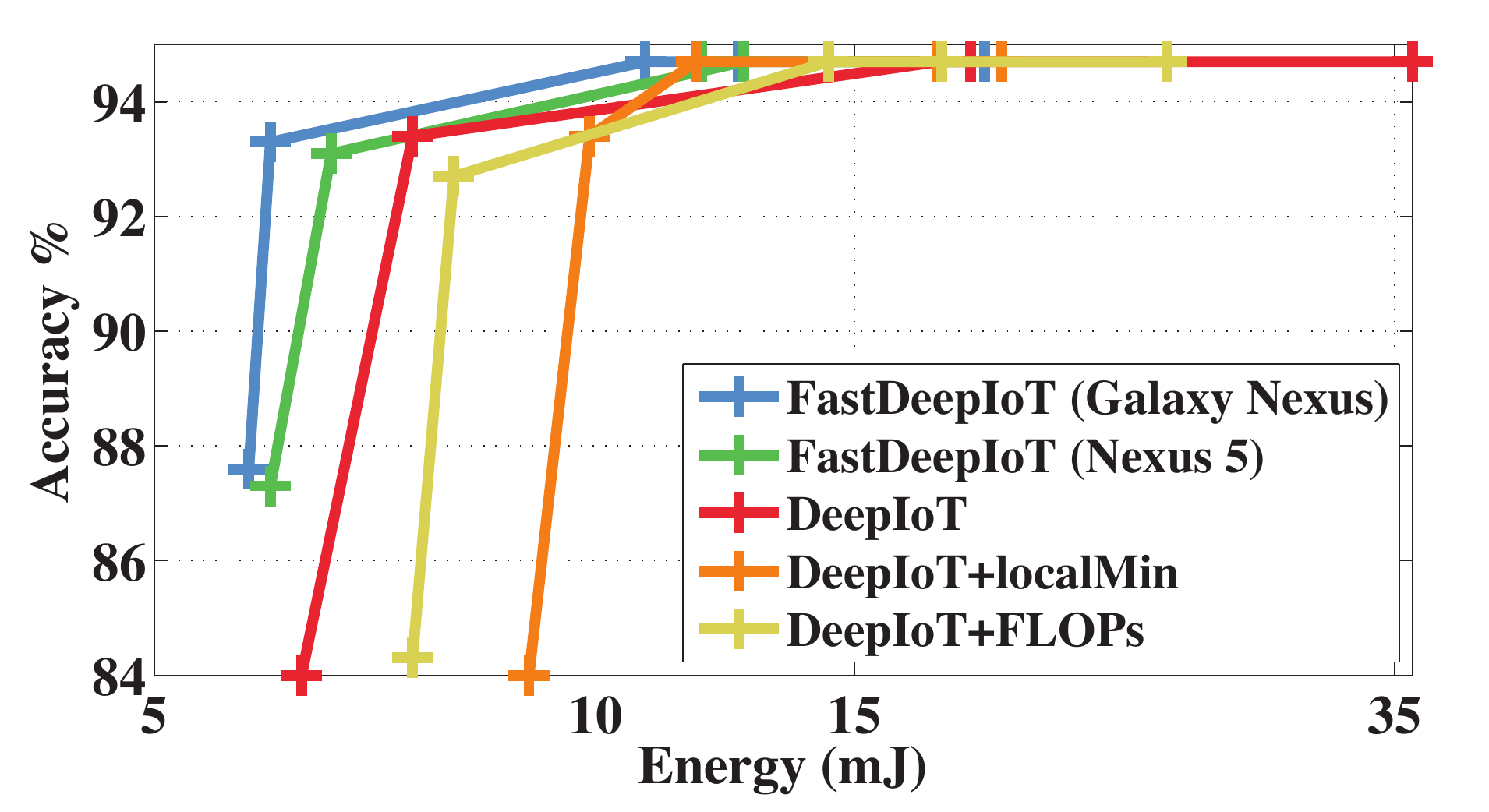}
  \vspace{-0.3cm}
  \caption{Tradeoff between testing accuracy and energy consumption on Galaxy Nexus.}
  \label{fig:HHAR_acc_energy_gn}
\end{subfigure}
\vspace{-0.2cm}
\caption{System performance tradeoff for DeepSense on HHAR dataset}
\label{fig:HHAR}
\vspace{-0.5cm}
\end{figure*}

Figure~\ref{fig:CIFAR10_acc_param} shows the tradeoff between testing accuracy and left proportion of model parameters. Since there is no algorithm targeting at minimizing model parameters, all methods show comparable performances. However, from another perspective, the execution time model learnt by FastDeepIoT empowers  existing compression algorithms to reduce more execution time with almost the same amount of parameters.

\vspace{-0.3cm}
\subsubsection{Large-scale image recognition on ImageNet}
This is a large-scale vision based task, image recognition based on a high-resolution camera. During this experiment, we use ImageNet as our training and testing dataset. The ImageNet dataset consists of 1.2 million $224\times224$ color images in 1000 classes with 100,000 images for testing. 

During the evaluation, we still use VGGNet structure as the original network structure. The detailed structures of best compressed models without accuracy degradation of all algorithms are shown in Table~\ref{tab:imagenet_vgg}. Note that the original VGGNet for $224\times224$ colour image input is too large for running on two testing hardware. FastDeepIoT achieves the best performance on the execution time among all methods. Compared with the state-of-the-art DeepIoT method, FastDeepIoT can further reduce the execution time by $59\%$ to $62\%$. DeepIoT+localMin still outperforms DeepIoT by reducing around $5\%$ to $10\%$ of execution time. In addition, FastDeepIoT can further reduce $25\%$ to $29\%$ of execution time compared with DeepIoT+FLOPs.

Figure~\ref{fig:imageNet_acc_time_n5} and~\ref{fig:imageNet_acc_time_gn}  shows the tradeoff between testing  top-5 accuracy and execution time for all algorithms. FastDeepIoT consistently outperforms all other algorithms by a significant margin. With the help of execution-time local minima, DeepIoT+localMin can still outperform DeepIoT in all cases. 
DeepIoT+FLOPs performs better than DeepIoT in this case. 

Figure~\ref{fig:imageNet_acc_energy_n5} and~\ref{fig:imageNet_acc_energy_gn} illustrates the tradeoff between testing top-5 accuracy and energy consumptions. FastDeepIoT outperforms all algorithms with a large margin. However, FastDeepIoT with the Galaxy Nexus execution time model is not the most energy-saving compression method on the Galaxy Nexus device. Also, DeepIoT+localMin cannot consistently outperforms DeepIoT on energy saving. These two observations witness the discrepancies between the execution time and energy modeling on mobile devices. Figure~\ref{fig:imageNet_acc_param} shows the tradeoff between testing accuracy and left proportion of model parameters. Again, all methods show the similar tradeoff, which indicates that FastDeepIoT is a parameter-efficient method on execution time reduction.

\subsubsection{Heterogeneous human activity recognition}~\label{sec:exp_HHAR}
This is a human-centric context sensing application, recognizing human activities with accelerometer and gyroscope. Especially, we are considering the heterogeneous human activity recognition (HHAR). This task focuses on the generalization ability with human who has not appeared in the training dataset. During this experiment, we use the dataset collected by Allan et al.~\cite{stisen2015smart}. During this evaluation, we use DeepSense structure as the original network structure~\cite{yao2017deepsense}. Table~\ref{tab:HHAR} illustrates the detailed structure of the original network and final compressed networks generated by four algorithms with no degradation on testing accuracy. As shown in Table ~\ref{tab:HHAR}, FastDeepIoT achieves the best performance on two devices with the corresponding execution time models. Compared with DeepIoT, FastDeepIoT can further reduce the model execution time by $22\%$ to $42\%$. During the compressing process, we observe that all compressed models tend to approach a model execution time lower bound, which has not been seen in the previous two experiments. In order to obtain the lower bound, we build a DeepSense structure with all hidden units that equal to 1, and then applies Algorithm~\ref{alg:local_minima} to find the structure that triggers local minimum. The resulted structure is illustrated in Table ~\ref{tab:HHAR} denoted by $t_\text{min}$. If we calculate the deductible model execution time by subtracting $t_\text{min}$ from the model execution time, compared with DeepIoT, FastDeepIoT can reduce the deductible execution time by $69\%$ to $78\%$. 

Furthermore, we can attempt to deduce the fundamental cause of the lower bound with our execution time model. As shown in \eqref{eqn:predictive_updated}, the execution time of recurrent layer is partially controlled by the number of \textit{step}, which can be interpreted as an initialization overhead for each \textit{step} in the recurrent layer. We can use an example to illustrate the relationship between the \textit{step} overhead and this lower bound. In our experiment, there are $20$ steps in the GRU. The coefficient of $step$ on Nexus 5 is $0.666$ ms. Therefore, the lower bound is $14.1\approx20\times 0.666$ ms. Thus, only algorithms dealing with reducing recurrent-layer steps can help further reducing the model execution time. 
Unfortunately, to the best of our knowledge, there is no existing work that solves this problem. However, our empirical observation and execution time model reveal  an interesting problem that requires future research.

The tradeoffs between testing accuracy and execution time for different algorithms are illustrated in Figure~\ref{fig:HHAR_acc_time_n5} and~\ref{fig:HHAR_acc_time_gn}. FastDeepIoT still achieves the best tradeoff for all cases. The tradeoffs between testing accuracy and energy consumption are illustrated in Figure~\ref{fig:HHAR_acc_energy_n5} and~\ref{fig:HHAR_acc_energy_gn}. FastDeepIoT performs better than all other baselines in almost all cases. The tradeoffs between testing accuracy and remanining proportion of model parameters are illustrated in Figure~\ref{fig:HHAR_acc_param}. All algorithms show comparable results.

\vspace{-0.4cm}
\section{Related Work}~\label{sec:related}
A key direction in embedded sensing literature is to speed up progressively more complex and interesting applications on resource-constraint embedded and mobile devices. 
Recent studies start focusing on speeding up deep neural networks through model compression. Han et al. propose a magnitude-based compression algorithm, illustrating promising results on resource-efficient deep neural networks with model compression~\cite{han2015deep}. Bhattacharya et al. design a sparse-coding and matrix factorization based solution to factorize neural networks into low-complexity structure for reducing resource consumption~\cite{bhattacharya2016sparsification}. Yao et al. propose a reinforcement learning based adaptive dropout solution to explore the less-redundant network structure for mobile and embedded devices~\cite{yao2017deepiot}. All these previous compression algorithms focus on reducing the model parameters, while taking execution time speed-up as a by-product. Therefore, these compression methods inevitably show inferior performance on execution time reduction. 
To the best of our knowledge, FastDeepIoT is the first framework to understand the impact of changing neural network structure on model execution time, and to empower existing compression algorithms to reduce the execution time on mobile and embedded devices properly.

\vspace{-0.35cm}
\section{Conclusion and Future Work}
\vspace{-0.2cm}~\label{sec:conclusion}
In this paper, we introduced FastDeepIoT, a framework for understanding and minimizing neural network execution time on mobile and embedded devices. 
We proposed a tree-structured linear regression model to figure out the causes of execution-time nonlinearity and to interpret execution time through explanatory variables. Furthermore, 
we utilized the execution time model to rebalance the focus of existing structure compression algorithms to reduce the overall execution time properly. We evaluated FastDeepIoT with three representative sensing tasks on two devices, where FastDeepIoT outperformed the state-of-the-art algorithms on reducing execution time and energy consumption with a large margin. 

This work is just a first step into the exploration of neural network compression for performance optimization. More profiling results are needed with the different choices of hardware, OS versions, load factors, power scaling, and deep learning libraries. 
Currently, FastDeepIoT can only support deep learning {\em structure\/} compression algorithms. More work is needed to support other deep learning compression methods, such as parameter quantization and pruning~\cite{han2015deep}.
The execution time model shows that the setup overhead of recurrent layers imposes a lower bound on efficacy of compression. It is a function of recurrent neural network steps, offering another dimension to compress for speeding up recurrent layers.
These insights offer avenues for future research on system performance oriented neural network compression for sensing applications. 

\vspace{-0.3cm}
\begin{acks}
\vspace{-0.15cm}
Research reported in this paper was sponsored in part by NSF under grants CNS 16-18627 and CNS 13-20209 and in part by the Army Research Laboratory under Cooperative Agreements W911NF-09-2-0053 and W911NF-17-2-0196. The views and conclusions contained in this document are those of the authors and should not be interpreted as representing the official policies, either expressed or implied, of the Army Research Laboratory, NSF, or the U.S. Government. The U.S. Government is authorized to reproduce and distribute reprints for Government purposes notwithstanding any copyright notation here on.
\end{acks}

}

\newpage
\bibliographystyle{abbrv}
\bibliography{reference}

\end{document}